\documentclass[10pt,a4paper]{article}
    \usepackage{longtable}

\usepackage{booktabs}
\usepackage{times}
\usepackage{url}
\usepackage{latexsym}
\usepackage{amsmath}
\usepackage{multirow}
\usepackage{natbib}
\usepackage{bbding}
\usepackage{pifont}
\usepackage{array}
\usepackage{graphicx,verbatim}
\usepackage{algorithm,algpseudocode}
\usepackage{float}
\usepackage{booktabs}

    \usepackage[T1]{fontenc}
    \usepackage{inputenc}
    \usepackage[english]{babel}
    \usepackage[compact,raggedright,small]{titlesec}
    \usepackage[affil-it]{authblk}
    \usepackage[runin]{abstract}
    \usepackage{multicol}
    \usepackage{blindtext}
    \usepackage{microtype}
    \usepackage{breakcites}

\newcommand{\checkmark}{\ding{51}}%
\newcommand{\xmark}{\ding{55}}%
    \addto{\Affilfont}{\small}

    \title{An Empirical Evaluation of Sequence-Tagging Trainers}
    \date{}
    \author[1]{P. Balamurugan}
    \author[1]{Shirish Shevade}
    \author[2]{S. Sundararajan}
    \author[3]{S. S. Keerthi}
    \affil[1]{Computer Science and Automation, Indian Institute of Science, Bangalore, India.}
    \affil[2]{Microsoft Research, Bangalore, India.}
    \affil[3]{Cloud and Information Services Lab, Microsoft Corporation, Mountain View, CA, USA.}

\begin{document}
      \maketitle

\begin{abstract}
The task of assigning label sequences to a set of observed sequences is common
in computational linguistics.
Several models for sequence labeling have been proposed
over the last few years. 
Here, we focus on
discriminative models for sequence labeling. Many batch and online (updating
model parameters after visiting each example) learning
algorithms have been proposed in the literature. On large
datasets, online algorithms are preferred as batch learning
methods are slow. These online algorithms were designed to
solve either a primal  or a dual problem. However, there has been no systematic
comparison of these algorithms in terms of their speed,
generalization performance (accuracy/likelihood) and their ability
to achieve steady state generalization performance fast. With this aim, we
compare different algorithms and make
recommendations, useful for a practitioner.
We conclude that the selection of an algorithm for sequence labeling depends on the  evaluation criterion used and its implementation simplicity.
\end{abstract}

\def\bw{{\bf {w}}}
\def\bx{{ \bf {x} }}
\def\by{{ \bf {y} }}
\def\byihat{\hat{\by}_i}
\def\byibar{\bar{\by}_i}
\def\byhat{\hat{\by}}
\def\bxi{{\mbox{\boldmath $\xi$}}}
\def\alphaiy{{\mbox{\boldmath $\alpha$}_i({\by})}}
\def\alphaiyq{{\mbox{\boldmath $\alpha$}_i({\by}_q)}}
\def\alphaiyp{{\mbox{\boldmath $\alpha$}_i({\by}_p)}}
\def\alphaiyi{{\mbox{\boldmath $\alpha$}_i({\by}_i)}}
\def\alphaiyhat{{\mbox{\boldmath $\alpha$}_i(\hat{\by}_i)}}
\def\delfiy{\Delta f_i ({\by})}
\def\delfiyq{\Delta f_i ({\by}_q)}
\def\delfiyp{\Delta f_i ({\by}_p)}
\def\delfiyihat{\Delta f_i (\hat{\by}_i)}
\def\fiyi{f({\bx_i},{\by_i})}
\def\fiy{f({\bx_i},{\by})}
\def\fiyhat{f({\bx_i},{\hat{\by}})}
\def\delfiybaryi{\Delta f_i (\bar{\by}_i)}
\def\Fiy{F_i ({\by})}
\def\Giy{G_i ({\by})}
\def\Alpha{{\mbox{\boldmath $\alpha$}}}
\def\Alphai{{\mbox{\boldmath $\alpha_i$}}}
\def\delAlphai{{\mbox{\boldmath $\delta \alpha_i$}}}
\def\delalphaiy{{\mbox{\boldmath $\delta \alpha$}_i({\by})}}
\def\delalphaiyp{{\mbox{\boldmath $\delta \alpha$}_i({\by}_p)}}
\def\delalphaiyq{{\mbox{\boldmath $\delta \alpha$}_i({\by}_q)}}
\def\half{{1 \over 2}}
\def\Y{\cal Y}
\def\X{\cal X}
\def\W{\cal W}
\def\liy{l_i({\by})}
\def\liybari{l_i(\bar{\by}_i)}
\def\dsp{\;\;}
\def\dow{\partial}
\def\l{\lambda}
\def\Hiy{H_i(\by)}
\def\Hiybari{H_i(\bar{\by}_i)}
\def\pstar{p^\ast}
\def\qstar{q^\ast}
\def\delalphaiypstar{{\mbox{\boldmath $\delta \alpha$}_i({\by}_{\pstar})}}
\def\delalphaiyqstar{{\mbox{\boldmath $\delta \alpha$}_i({\by}_{\qstar})}}
\def\alphaiyqstar{{\mbox{\boldmath $\alpha$}_i({\by}_{\qstar})}}
\def\alphaiypstar{{\mbox{\boldmath $\alpha$}_i({\by}_{\pstar})}}
\def\delfiyqstar{\Delta f_i ({\by}_{\qstar})}
\def\delfiypstar{\Delta f_i ({\by}_{\pstar})}
\def\mycite#1{(\citeauthor{#1}, \citeyear{#1})}
\def\myshortcite#1{\citeauthor{#1} (\citeyear{#1})}

\input{amssymb.mac}

\section{Introduction}
\label{intro}
In conventional classification learning, the aim is to learn a function which assigns
discrete (scalar) labels to unseen objects, given a set of already labeled training set
examples. There exist tasks in computational linguistics or bioinformatics, which are
often described as mappings from input sequences to output sequences. As an example,
in computational linguistics, such tasks include part-of-speech (POS) tagging, named
entity recognition (NER) and shallow parsing~\mycite{lbfgs}.

In this work, we focus on discriminative models for sequence learning. 
\myshortcite{crf} introduced conditional random fields (CRFs), an undirected graphical
model that models $p({\by}|{\bx})$ directly, and proposed to use iterative scaling algorithms
for CRF training. Subsequently, ~\myshortcite{lbfgs} demonstrated that preconditioned
conjugate gradient or limited-memory quasi-Newton (L-BFGS) methods offer significant
training speed advantages over iterative scaling. These batch algorithms were found to
be very slow on large sequence labeling problems. On a benchmark data set with about
$35,000$ examples and $18$ million features, L-BFGS method for CRF training required about
\emph{five} days on a reasonably fast machine to design a classifier. Therefore, in this
work, we consider online algorithms where model parameters are updated after
visiting each example.
Note that, though some of the
algorithms discussed here are fundamentally batch algorithms, they have an online feel
and are reasonably fast.

Many real world prediction problems can be posed as structured prediction problems,
where the output is a structured object like a sequence or a tree or a graph.
Large margin methods like support vector machines (SVMs) have shown much
promise for structured output learning~\mycite{tsochan}.
In recent years, several learning algorithms have been proposed to solve the 
structured prediction problems involving sequence
labeling. Some prominent methods among them include stochastic gradient descent (SGD)
algorithm for CRFs~\mycite{sgd}, Sequential Minimal Optimization (SMO)~\mycite{m3n}, 
Cutting Plane method~\mycite{Joa2009},
Sequential Dual Method (SDM)~\mycite{sdm}, exponentiated gradient (EG) methods~\mycite{eg}  
for max-margin Markov networks (also called Structural SVMs) 
and
structured perceptron~\mycite{spcollins}. 
These methods assume exact inference, 
which is often computationally expensive. 
\myshortcite{Huang} 
proposed the variants 
of perceptron, 
called ``violation-fixing perceptrons", 
which use a violation (approximate inference) in each update.
It is important to note that all these algorithms
(except the structured perceptron algorithm) solve either a primal problem or a dual problem.
It will be therefore interesting to compare these algorithms in terms of their speed
and generalization performance achieved by the resulting model. 
\myshortcite{nguyen}  compared
some prominent algorithms for sequence labeling. However, as pointed out in~\mycite{crfcomp},
this comparison employed different internal feature functions and therefore the comparison was not fair. In the case of sequence labeling, to the best of our knowledge,
there has been no systematic comparison of models obtained by using different algorithms, which solve either a primal
or a dual problem and use the same set of feature functions. We believe that, this evaluation will be useful for practitioners and help
them choose an appropriate method for sequence labeling depending upon the requirement.

{\bf Contributions}: 
This work is motivated by the need to compare different sequence labeling
algorithms systematically on real-world data sets and make recommendations about the algorithm selection.
We consider two types of convex loss functions and compare methods which solve the
regularized loss minimization problem (either the primal problem or its equivalent dual).
The loss functions used are: a) A variant of the hinge
loss function used in large-margin classification problems, and b) the negative 
log-likelihood function (used in CRFs).
In particular, we compare i) stochastic gradient descent (SGD) method for CRF to solve 
the primal problem for regularized loss function (b), ii) Cutting Plane (CP) method
to solve the dual problem of regularized loss function (a), iii) Sequential Dual Method (SDM) 
(which can be used to solve the dual problems obtained using either of the loss functions), and
iv) Averaged Structured Perceptron Algorithm (which does not use any objective function).
On a number of experiments carried out on large-scale real-world data sets, we observed that
the sequential dual method for SVMs and stochastic gradient descent methods should be preferred if test set accuracy and likelihood are respectively the evaluation criteria.
Further, the averaged structured perceptron algorithm does achieve comparable 
test set accuracy, if not test set likelihood. This is despite the fact that it
does not optimize any objective function! 

The paper is organized as follows. 
The following section briefly describes 
different sequence learning algorithms used in this work.
The details of various experiments 
performed on large data sets and their results 
are 
given in Section~\ref{experiments}. 
Our recommendations are presented in Section~\ref{conclusion}.

\section{Sequence Learning Algorithms}
\label{seqla}
In this section, we describe various sequence learning algorithms used in this work.
We assume that a training set $S$ of input-output sequence pairs is available. Let $S = \{ {\bx}_i, {\by}_i \}^n_{i=1}$ where ${\bx}_i \in \X$
and ${\by}_i \in \Y$ for every $i$. The goal is to learn  a discriminant
function $g:\X \times \Y \rightarrow \rr$ over the training set $S$
from which prediction for a input ${\bx}$ is given by
$$
h({\bx}) = \arg\max_{{\by} \in \Y} \;\;g({\bx}, {\by}).
$$
For sequence labeling problems, the $\arg\max$ computation in the above equation can
be done using dynamic programming like the Viterbi algorithm.
If an input sequence and the corresponding output sequence are of length $L$, and each individual
label of the output sequence can take values from the set $\Sigma$, then the sequence labeling
problem can be considered as a multi-class classification problem with $|\Sigma|^L$ classes. This
demonstrates that the cardinality of $\Y$ grows exponentially with the size of ${\bx}$. In this
work, we assume that $g({\bx}, {\by})$ takes the form of a linear function,
$$
g({\bx}, {\by}) = {\bw}^T f({\bx}, {\by})
$$
where ${\bw}$ is a parameter vector and $f({\bx}, {\by})$ is a feature vector
relating input ${\bx}$ and ${\by}$. Note that the feature vectors $f({\bx},{\by})$
have a crucial effect on the performance of the designed structured classifier~\mycite{crfcomp}.

Using a variant of the hinge loss function, the sequence learning problem can be posed as an
extension of multi-class SVM problem as follows~\mycite{crammer}:
\begin{eqnarray}
\min_{{\bw}, \bxi}  \dsp  \frac{\lambda}{2} {\| {\bw} \|}^2 + \sum_i \xi_i \nonumber   \\
{\rm s.t.} \dsp  {\bw}^T {\delfiy} \ge \liy - \xi_i \;\; \forall \;i, {\by}
\label{eq:mm-primal}
\end{eqnarray}
where  $\delfiy = f({\bx}_i, {\by}_i) - f({\bx}_i, {\by})$ and  $\lambda > 0$ is a regularization parameter. $l_i({\by})$ in~(\ref{eq:mm-primal}) is a loss function that quantifies the loss associated with predicting ${\by}$ instead of the correct output ${\by}_i$. In sequence learning problems,
a natural choice for the loss function is the Hamming distance,
$$
l_i({\by}) = \sum^L_{j=1} I(y^j_i \ne y^j)
$$
where $I(\cdot) $ is the indicator function and ${\by} = (y^1,y^2,\ldots,y^L)$.
By defining the
conditional distribution,
\begin{eqnarray}
p({\by} | {\bx}; {\bw}) = \frac{e^{{\bw}^T f({\bx}, {\by})}}{Z_x}
\label{eq:pd}
\end{eqnarray}
where $Z_x = \sum_{y^\prime} e^{{\bw}^T f({\bx}, {\by}^\prime)}$ is the partition function
and using the negative log-likelihood function, the parameter ${\bw}$ can be learned
by solving the following unconstrained optimization problem:
\begin{eqnarray}
\min_{{\bw}}  \dsp  \frac{\lambda}{2} {\| {\bw} \|}^2 - \sum_i \log p({\by}_i | {\bx}_i; {\bw})   
\label{eq:ll-primal}
\end{eqnarray}
where $\lambda > 0$ is a regularization parameter. CRF training (batch algorithm) involves
solving the problem~(\ref{eq:ll-primal}) in batch mode.
Note that both the problems~(\ref{eq:mm-primal}) and~(\ref{eq:ll-primal}) are convex programming
problems and the optimal solution ${\bw}$ can be used for
making the prediction for a specific input ${\bx}$ as 
$\arg\max_{{\by} \in \Y} \;\;{\bw}^T f({\bx},{\by})$. 

We note that a variety of techniques
have been developed in the literature to solve large scale sequence learning
problems. These include bundle method~\mycite{bundle}, fast Newton-CG method for batch learning of CRFs~\mycite{newtoncgcrf}, 
SGD and block coordinate methods for 
$L_1$ regularized and elastic-net CRFs~\mycite{practicalcrf}, 
stochastic 
meta-descent method~\mycite{crfsgdaccel}, 
stochastic block coordinate Frank-Wolfe optimization~\mycite{frankwolf}, 
dual coordinate ascent method~\mycite{gimpelsmith} and 
excessive gap reduction technique~\mycite{egspeedup}.   
The number of dual variables for the problems~(\ref{eq:mm-primal}) and~(\ref{eq:ll-primal}) is exponentially large and techniques which use 
marginal variables (which are polynomial in number and from which dual 
variables can be easily derived) were also
proposed in the literature. 
These techniques include exponentiated gradient method~\mycite{eg} and
sequential minimal optimization algorithm~\mycite{m3n}.  
It will be difficult to 
compare every method proposed in the literature. Therefore, in this work, we 
restrict ourselves to algorithms which are simple, easy to implement 
from a practitioner's viewpoint and solve
the problems~(\ref{eq:mm-primal}) and~(\ref{eq:ll-primal}) or their dual 
problems directly. 

Recently, adaptive-rate and parameter-free variants 
of SGD have also been proposed 
for binary classification tasks in 
\mycite{adaptivesgd} and \mycite{peskysgd}.
While a simple SGD chooses the learning rate 
to be $\frac{1}{t}$, $t$ being the number 
of iterations, 
adaptive-rate variant of SGD \mycite{adaptivesgd} 
updates the learning rate for each component of 
the model parameter ${\bw}$ by incorporating 
gradient-information from the past iterations. 
However, tuning is required to find  
an initial choice of the 
learning rate. 
\myshortcite{peskysgd} propose to 
automatically choose the learning rate based on 
a second-order approximation of the loss function 
at a particular component of $\bw$, 
thus eliminating the need to tune the learning 
rate altogether. 
However, making the learning rate 
automatically-tuned or adaptable,  
comes at an added cost of 
computing the second order information 
or book-keeping gradient information, thus 
increasing the iteration complexity of the basic SGD. 
Moreover, the results given in \mycite{peskysgd} 
show that the generalization performance  
achieved by the basic SGD is 
comparable and sometimes better than that 
achieved by the 
adaptive-rate and parameter-free variants 
of SGD. Hence, we resort to the basic SGD method with 
averaging~\mycite{sgd}, in our experiments. 

In the following, we give a brief description of 
the algorithms used to solve the problems~(\ref{eq:mm-primal}) and~(\ref{eq:ll-primal}) and compared in this work. 

\subsection{Cutting Plane Method ({\bf CP})}

\myshortcite{Joa2009} proposed an equivalent formulation of the problem 
~(\ref{eq:mm-primal}), given by~(\ref{eq:Joach}) 
and presented a cutting-plane algorithm, which is significantly fast on large scale problems. 
It was shown
that the dual problem of~(\ref{eq:Joach}) has a sparse solution (that is, the number of non-zero dual variables
is small at optimality). 
\begin{align}
\min_{{\bw}, \xi} \dsp  \frac{\lambda}{2} {\| {\bw} \|}^2 +  \xi \nonumber \\
{\rm s.t.} \dsp  \frac{1}{n} {\bw}^T \sum^n_{i=1} {\delfiy} \ge \frac{1}{n} \sum^n_{i=1} \liy - \xi, \nonumber \\ 
\; \forall \; \{{\by}_1,\ldots,{\by}_n\} \in {\Y}^n \label{eq:Joach}
\end{align}
Even for large data sets, the size of the quadratic programs that need to be solved
was observed to be very small (as the number of violated constraints was very small) and therefore, the method achieved considerable speed-up.
We note that the bundle method presented in~\mycite{bundle} is similar to
the cutting-plane algorithm and therefore, we do not use it for comparison.

\subsection{Sequential Dual Method for Structural SVMs ({\bf SVM-SDM})}

\myshortcite{sdm} suggested the use of sequential dual method (SDM) to solve the dual problem of~(\ref{eq:mm-primal}).
\begin{align}
\min & \frac{1}{2 \lambda} {\| \sum_{i, {\by}} \alphaiy \delfiy \|}^2 - \sum_{i, {\by}} \alphaiy \liy \nonumber \\
{\rm s.t.} & \sum_{\by} \alphaiy = 1 \; \forall \; i, \; \alphaiy \ge 0 \; \forall \; i, {\by} \label{eq:mm-dual}
\end{align}
The method makes repeated passes over the training data set and optimizes
the dual variables associated with one example at a time, until some
stopping condition is satisfied.
Note that the number of dual variables in~(\ref{eq:mm-dual}) is exponentially large for every example $i$.
However, at optimality, very few of the dual variables are strictly positive. 
This fact was used to develop a fast and
efficient algorithm to solve the dual problem~(\ref{eq:mm-dual}). 
Some heuristics were also proposed to make the sequential dual method more
efficient.
This method was found to
be an order of magnitude faster than the cutting-plane method on many sequence-learning data sets~\mycite{sdm}.

\subsection{Averaged Stochastic Gradient Descent Method for CRF ({\bf CRF-ASGD})}

Gradient based online methods like stochastic gradient descent (SGD) can be used to solve the CRF primal
problem~(\ref{eq:ll-primal})~\mycite{sgd}. The SGD method is fast and is quite useful when the training data
size is large. It operates by visiting each example and updating the parameter ${\bw}$ through a simple update
step. For example, at iteration $t$, using a single training example (say ${\bx}_i$), the following simple
update rule is used:
\begin{align}
{\bw}_{t+1} = {\bw}_t - \gamma_t (\lambda {\bw}_t + f({\bx}_i, {\by}_i) -  E_{p(y|x_i)} f({\bx}_i, {\by})) \\ \nonumber
\end{align}
where $\gamma_t$ is a learning rate parameter (typically set to $\frac{1}{t+1}$). The expectation term,
$E_{p(y|x_i)} f({\bx}_i, {\by})$ is calculated with respect to the conditional probability~(\ref{eq:pd}).
This is usually done using a forward-backward algorithm~\mycite{crf}. The SGD method, though simple, requires multiple
passes over the data before it converges to the optimal solution. To overcome this difficulty, ~\myshortcite{asgd}
proposed a method which averages the parameter ${\bw}$. An average parameter $\bar{\bw}$ is maintained and updated
at every iteration,
\begin{eqnarray}
\bar{{\bw}}_{t+1} = \frac{t}{t+1} \bar{\bw}_t + \frac{1}{t+1} {\bw}_{t+1}. 
\end{eqnarray}
This method, called averaged SGD (ASGD), has been demonstrated to make reasonable progress
in the objective function in the initial few iterations. A thorny issue with online
methods like SGD or ASGD methods is the choice of the initial learning rate, $\gamma_0$. With
an improper choice, the methods might become very slow on large data sets. Choosing a suitable
$\gamma_0$ involves taking a random sample of the data set initially and performing SGD or ASGD
updates on this selected sample using different learning rate values and then choosing the best
learning rate as the rate which gives maximum decrease in the objective function. Having determined
the initial learning rate $\gamma_0$, the method then proceeds in the normal fashion on the entire
training set.

\subsection{ Sequential Dual Method for CRF ({\bf CRF-SDM})}

Inspired by the speed-up achieved by SVM-SDM over state-of-the-art methods, we implemented a sequential dual
method to solve the dual of the CRF primal problem~(\ref{eq:ll-primal}). 
\myshortcite{memisevic} and \myshortcite{maxent} proposed to
solve the dual of~(\ref{eq:ll-primal}) for multi-class classification problems. These methods
cannot be directly used for sequence learning problems as the number of dual variables is exponentially
large for such problems. Further, the resulting model at optimality is not sparse. 
That is, all the possible sequences are used to define the parameter vector, 
\begin{align}
\bw (\Alpha)=\frac{1}{\lambda} \sum_{i, {\by}} \alphaiy \delfiy \label{eq:wdual}
\end{align}
when the following dual problem of~(\ref{eq:ll-primal}) 
is solved: 
\begin{align}
\min \dsp \frac{1}{2 \lambda} {\| \bw (\Alpha) \|}^2 + \sum_{i, {\by}} \alphaiy \log \alphaiy \nonumber \\
{\rm s.t.} \dsp \sum_{\by} \alphaiy = 1 \; \forall \; i, \; \alphaiy \ge 0 \; \forall \; i, {\by} \label{crfdual}
\end{align}
One way to alleviate this problem 
is to assume that $\alphaiy \leq \eta$ for many $\by \in \Y$ corresponding to 
every example $i$, where $\eta$ is very small (say, $10^{-18}$) 
and solve the dual problem with 
respect to the sequences in the set, $V_i=\{ \by : \alphaiy > \eta\}$ 
using SMO-type algorithm with the 
modified Newton  method illustrated in \mycite{maxent}. 
Note that the CRF-SDM method finds only an \emph{approximate} 
solution to~\eqref{crfdual} because of the practical 
limitation on the size of the set $V_i$. 
Our experiments presented in the next section, 
clearly indicate that on many data sets, 
such an approximate solution is sufficient to obtain 
a comparable generalization performance. 
 
\subsection{Averaged Structured Perceptron ({\bf AvStructPerc})}

Perceptron algorithm for binary classification is simple and does not use any objective function.
\myshortcite{spcollins} proposed the perceptron algorithm for structural learning. After randomly initializing the
weight vector ${\bw}$, the algorithm makes repeated passes over the training data set (visiting one
example at a time), until some stopping condition is satisfied. In every pass, after visiting an example
$({\bx}_i, {\by}_i)$, the following update rule is used if the current weight vector fails to predict
the desired label ${\by}_i$:
\begin{eqnarray}
{\bw} := {\bw} + \eta (f({\bx}_i, {\by}_i) - f({\bx}_i, {\by}^\prime))
\end{eqnarray}
where ${\by}^\prime = \arg\max_y  \; {\bw}^T f({\bx}_i, {\by})$ and $\eta (> 0)$ is the learning rate parameter.
\myshortcite{spcollins} also proposed a simple refinement to the perceptron algorithm. Defining $w^{t,i}_j$ to be the
value of the $j$-th parameter after the $i$-th training example has been visited in pass $t$ over the
training data, the average parameter ${\bw}_{avg}$ is defined as,
\begin{eqnarray}
{\bw}_{{avg},j} = \sum^T_{t=1} \sum^n_{i=1} w^{t,i}_j / (nT) \;\; \forall \; j=1,\ldots,d
\end{eqnarray}
where $d$ denotes the dimension of the parameter vector ${\bw}$. 
In our implementation, we adopted a very efficient way of updating ${\bw}_{avg}$.
It was observed that the use of
this refined algorithm gave better generalization performance~\mycite{spcollins}.

 In our experiments, we observed
that the value of $\eta$ is crucial to get good generalization performance and therefore, preferred to
choose it using the procedure similar to that used in the SGD method. Since the averaged structured perceptron algorithm
does not optimize any objective function, the learning rate which resulted in the least error on the remaining
training set was chosen.

     \begin{table*}[ht]
    \caption{\textbf{Summary of data sets}. $n$, $n_{val}$ and $n_{test}$ denote the sizes of the training, validation and test data respectively, 
		  $d$ is the input dimension, $k$ denotes the number of alphabets and $N$ is the feature vector dimension}
    \label{dataset_tab} 
    \begin{center}
    \begin{tabular}{|p{20mm}|p{11mm}|p{11mm}|p{11mm}|p{15mm}|p{5mm}|p{15mm}|} 
    \hline %
    Data set & $n$ & $n_{val}$ & $n_{test}$ & $d$ & $k$ & $N$ \\\hline
    BioCreative & 6000 & 1500 & 5000 & 102409 & 3 & 307236 \\
    BioNLP & 14836 & 3710 & 3856 & 513932 & 11 & 5653373 \\
    CoNLL & 14987 & 3684 & 3466 & 651041 & 8 & 5208392\\
    dataPOS & 39832 & 2416 & 1700 & 258299 & 45 & 11625480\\
    WSJPOS & 28424 & 7107 & 1681 & 446147 & 42 & 18739938 \\\hline
    \end{tabular}  
    \end{center}
    \vspace{-.1in}
    \end{table*}

We give details 
of experimental evaluation in the next section. 

\section{Experiments and Discussion}
\label{experiments}
\begin{figure*}[ht]
\begin{minipage}[b]{0.5\linewidth}
\centering
\includegraphics[scale=0.4]{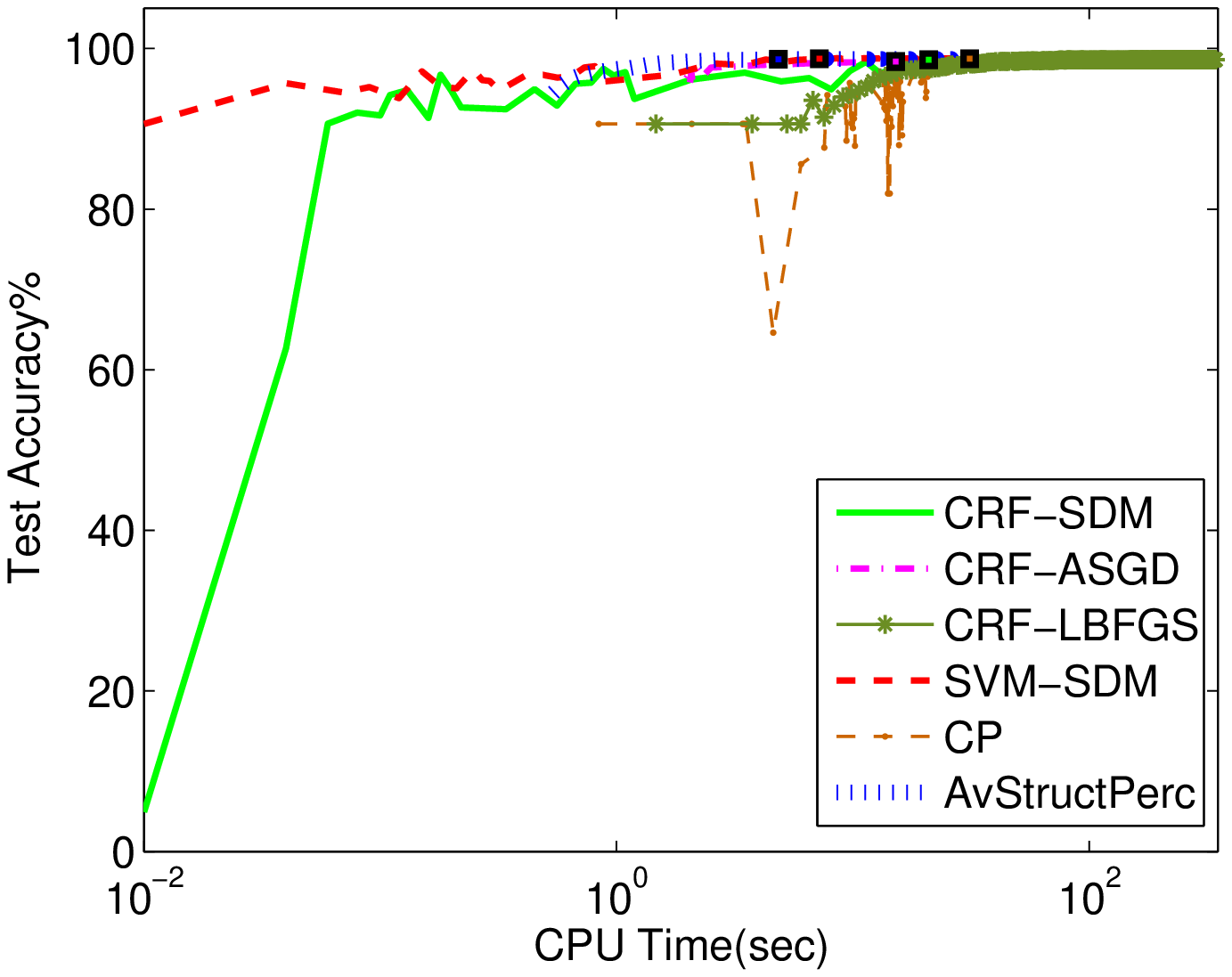}
\end{minipage}
\begin{minipage}[b]{0.5\linewidth}
\centering
\includegraphics[scale=0.4]{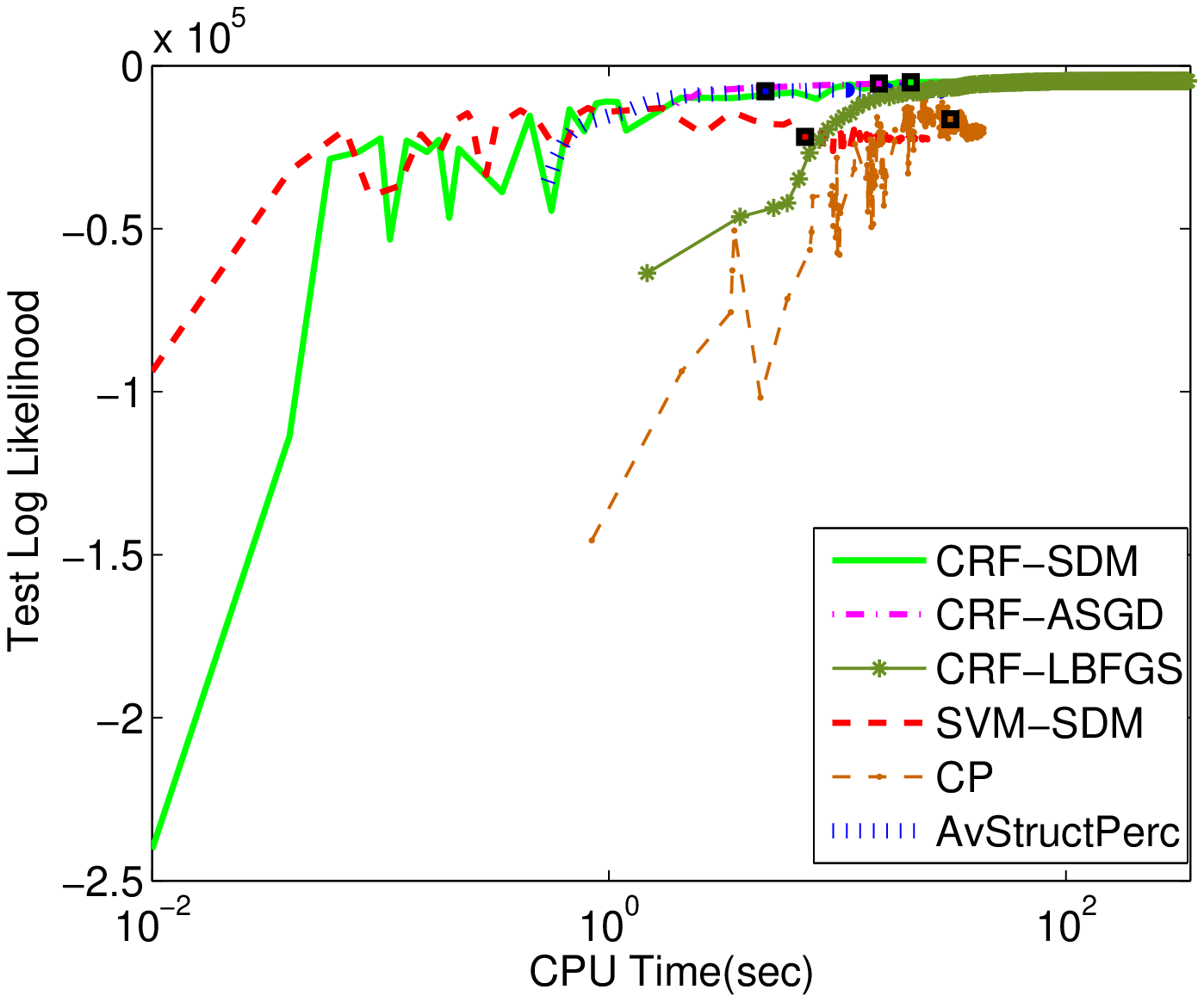}
\end{minipage}
\begin{minipage}[b]{0.5\linewidth}
\centering
\includegraphics[scale=0.4]{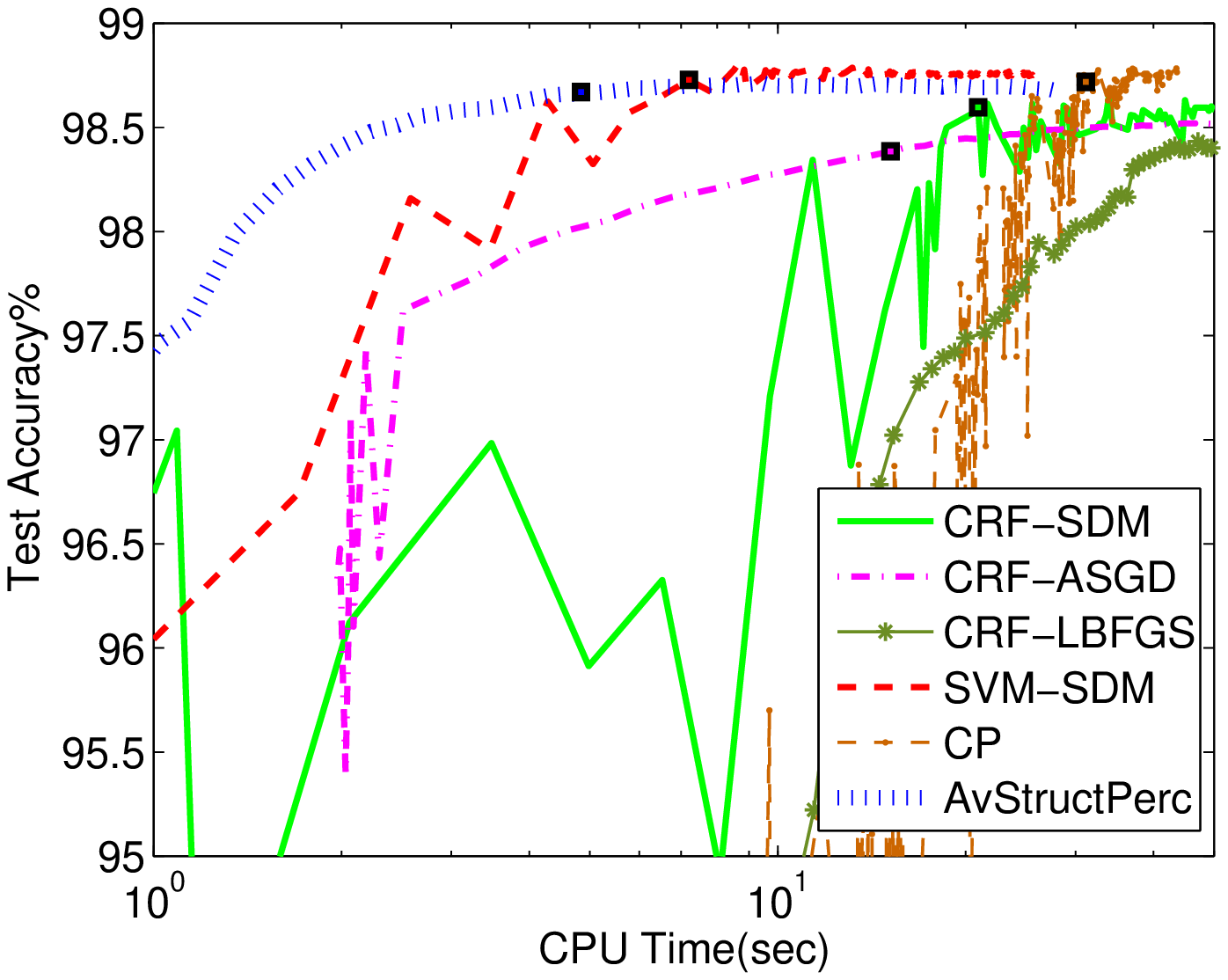}
\end{minipage}
\begin{minipage}[b]{0.5\linewidth}
\centering
\includegraphics[scale=0.4]{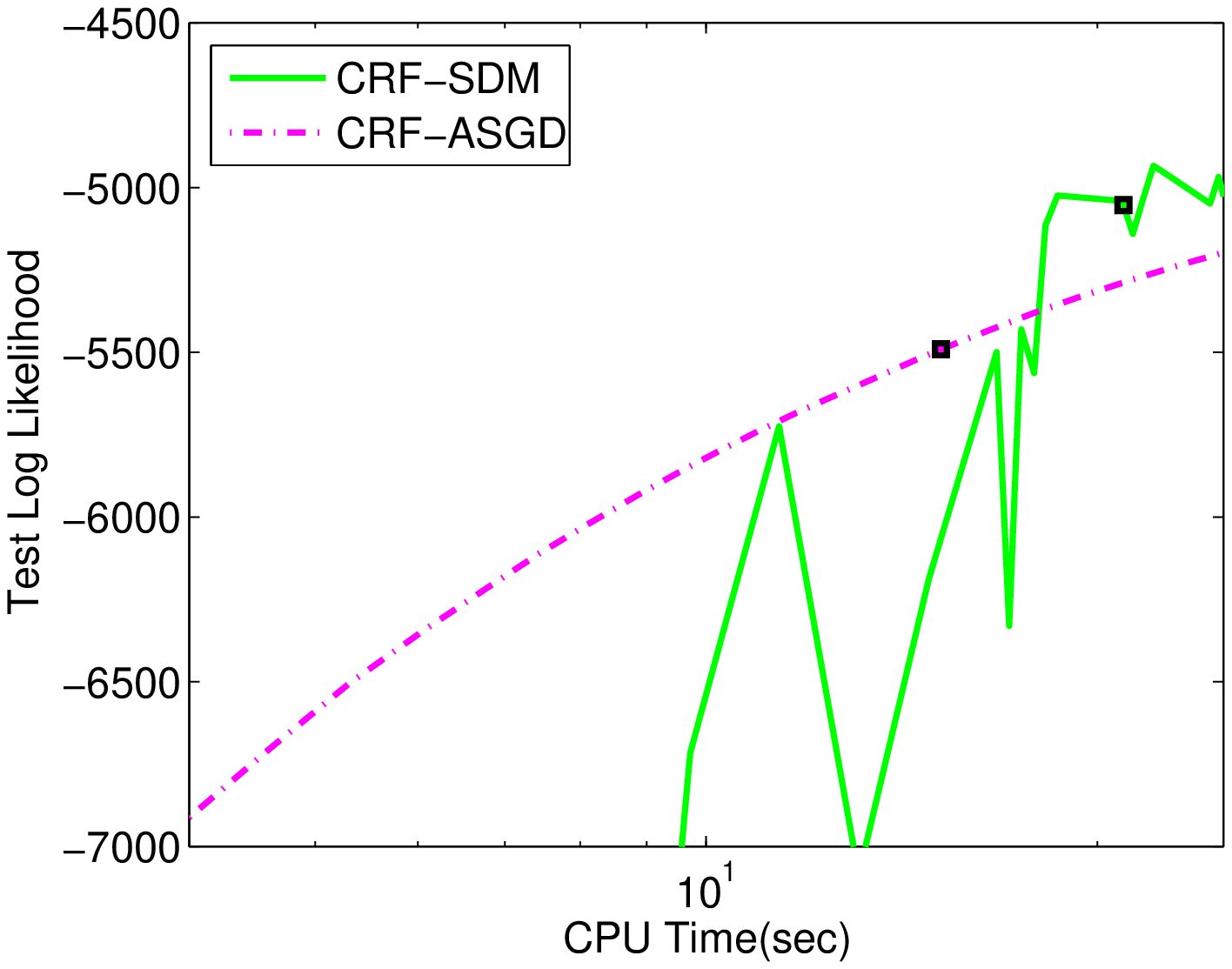}
\end{minipage}
\begin{minipage}[b]{0.5\linewidth}
\centering
\includegraphics[scale=0.4]{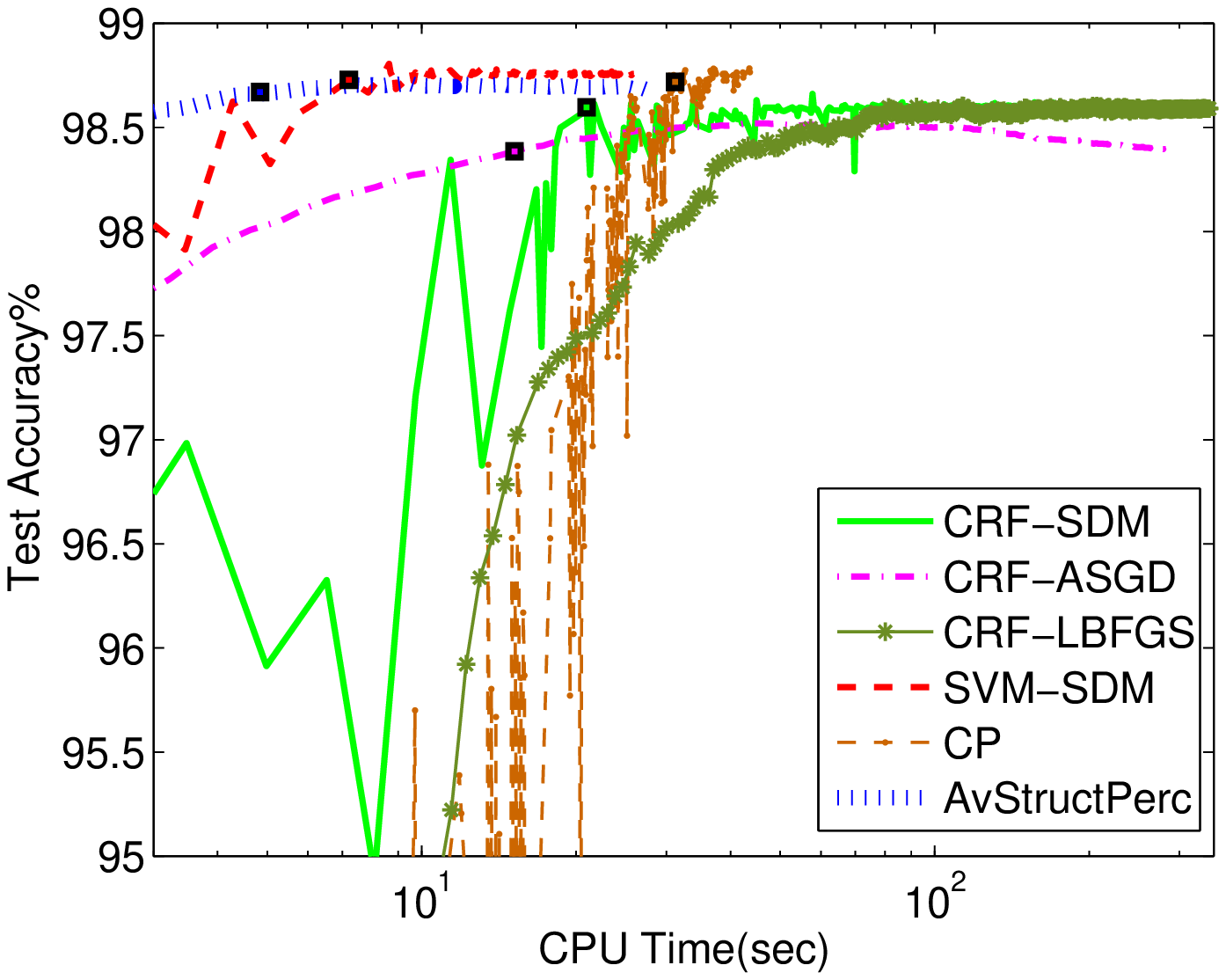}
\end{minipage}
\begin{minipage}[b]{0.5\linewidth}
\centering
\includegraphics[scale=0.4]{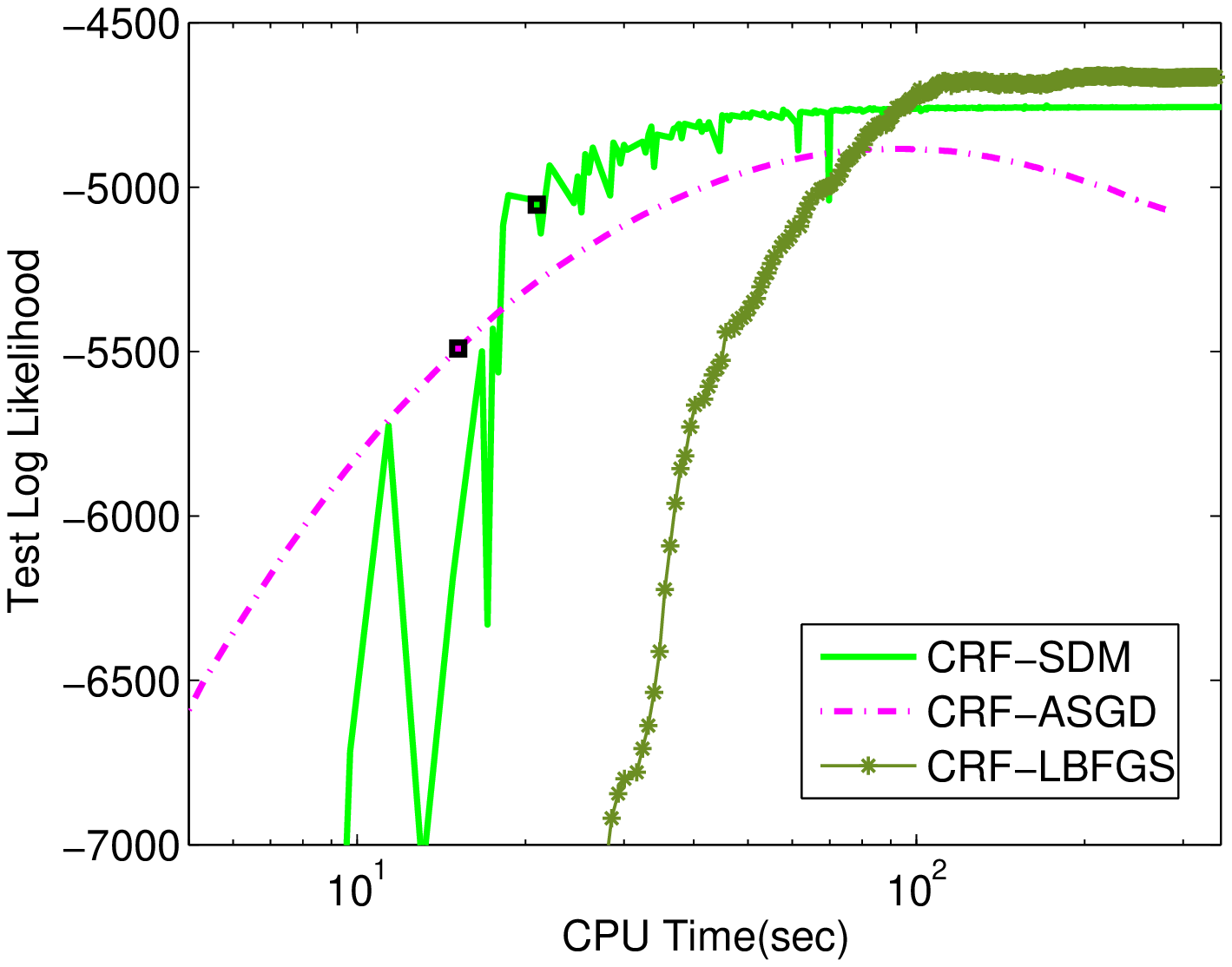}
\end{minipage}
\caption{\textbf{Comparison of Test Accuracy and Test Likelihood for BioCreative dataset.
 The plots in rows 2 and 3 are zoomed versions to clearly see certain
behaviour in the initial and final stages respectively.
} }
\label{biocreativefig}
\end{figure*}

\begin{figure}[!h]
\begin{minipage}[b]{0.5\linewidth}
\centering
\includegraphics[scale=0.4]{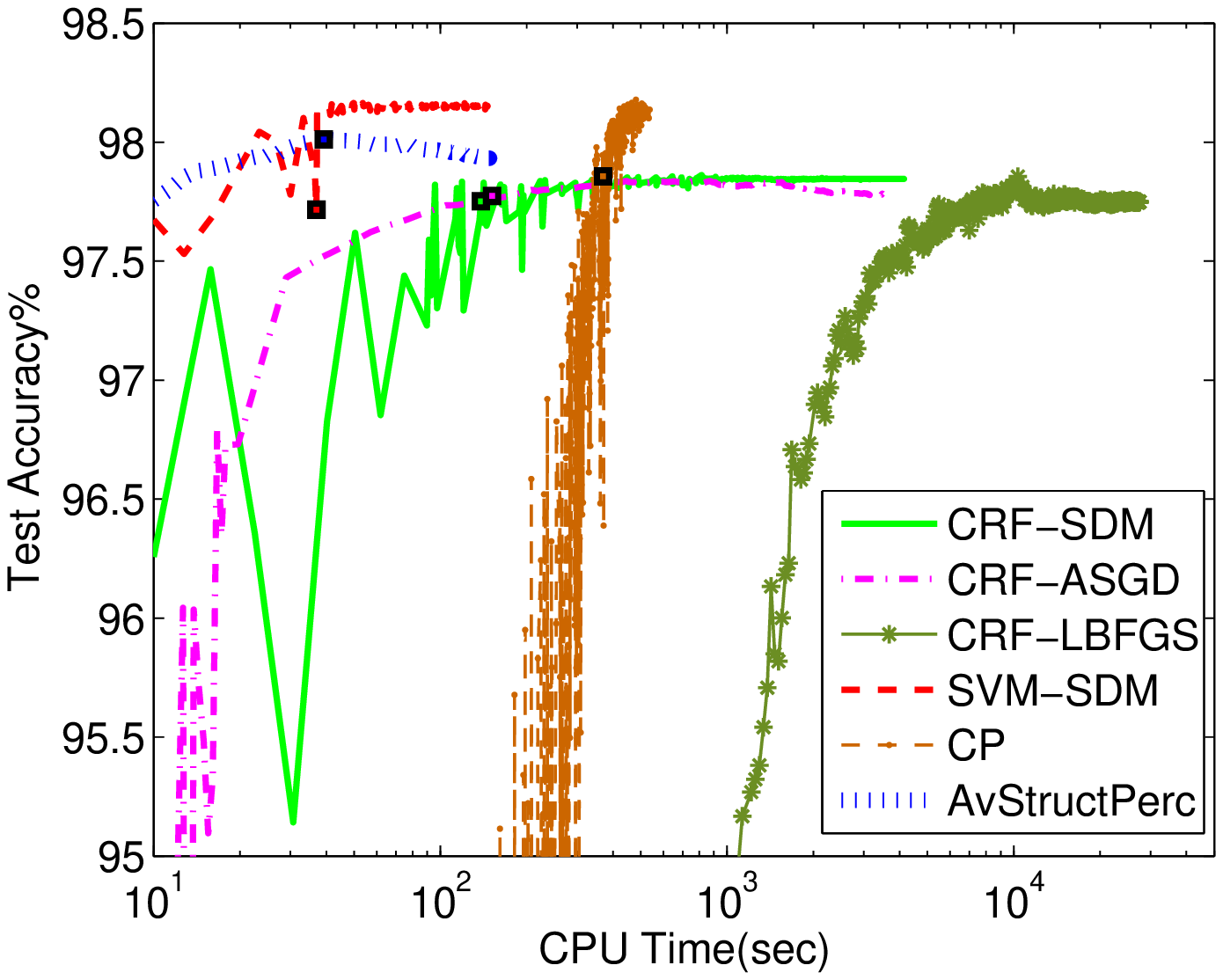}
\end{minipage}
\begin{minipage}[b]{0.5\linewidth}
\centering
\includegraphics[scale=0.4]{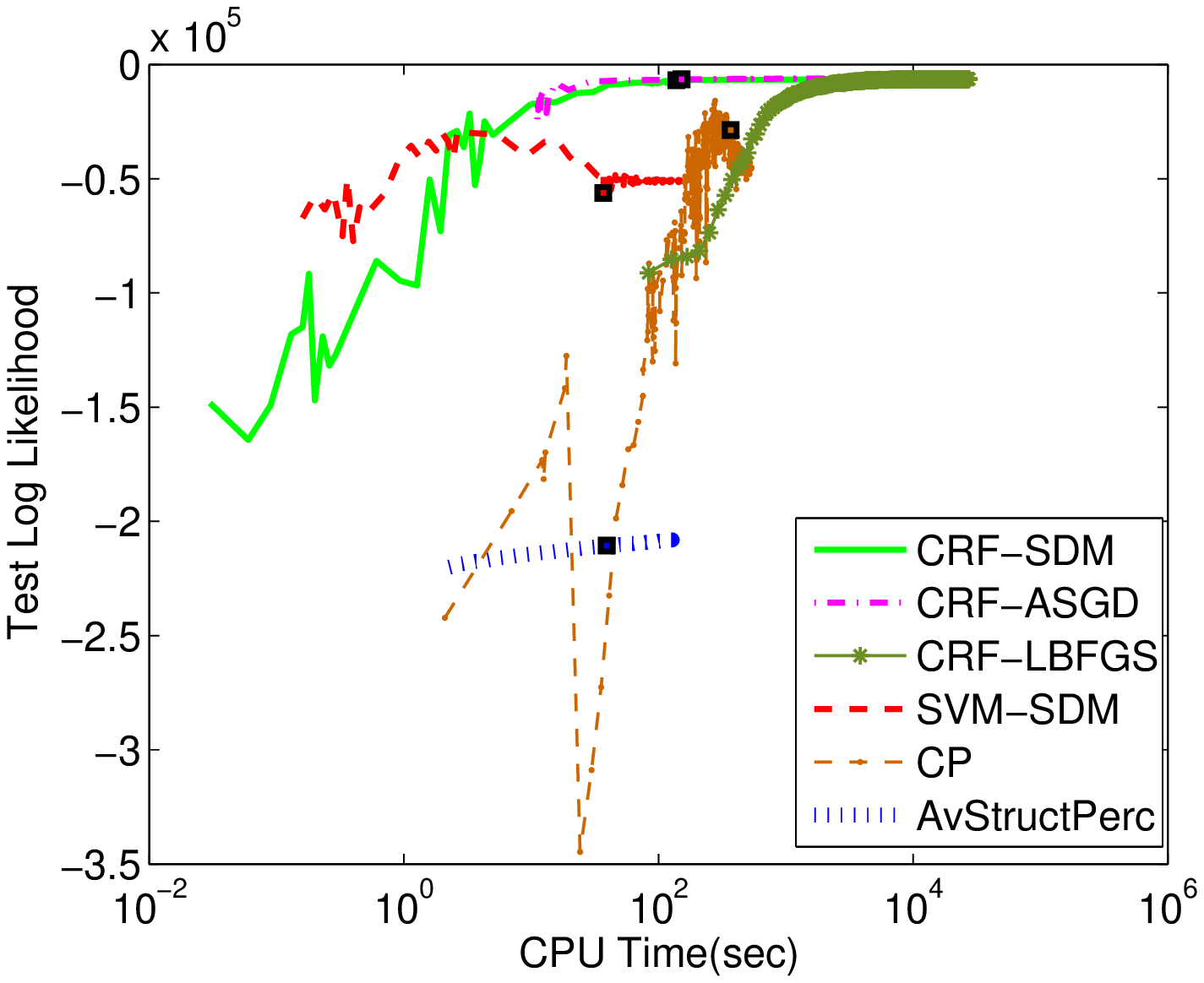}
\end{minipage}
\begin{minipage}[b]{0.5\linewidth}
\centering
\includegraphics[scale=0.4]{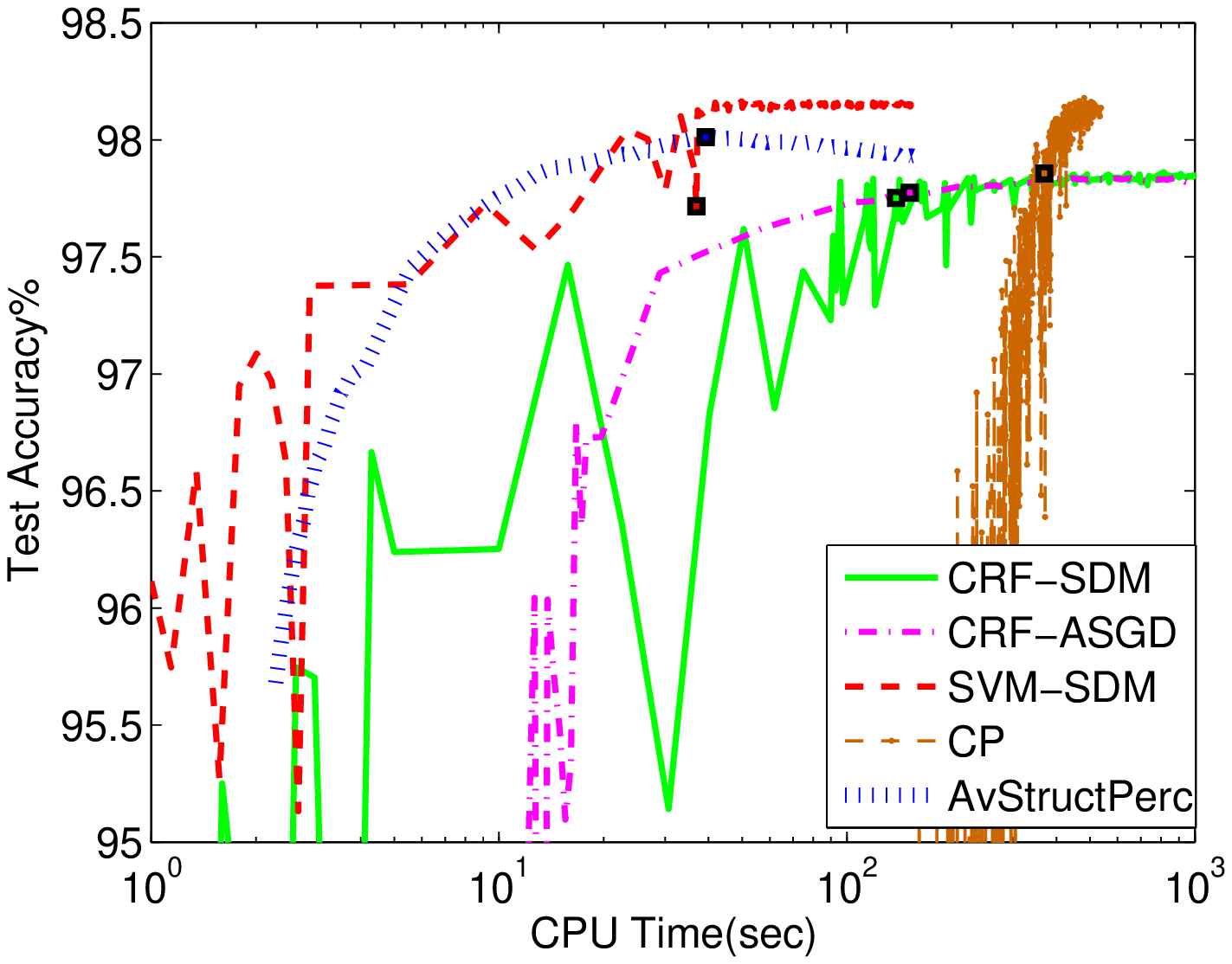}
\end{minipage}
\begin{minipage}[b]{0.5\linewidth}
\centering
\includegraphics[scale=0.4]{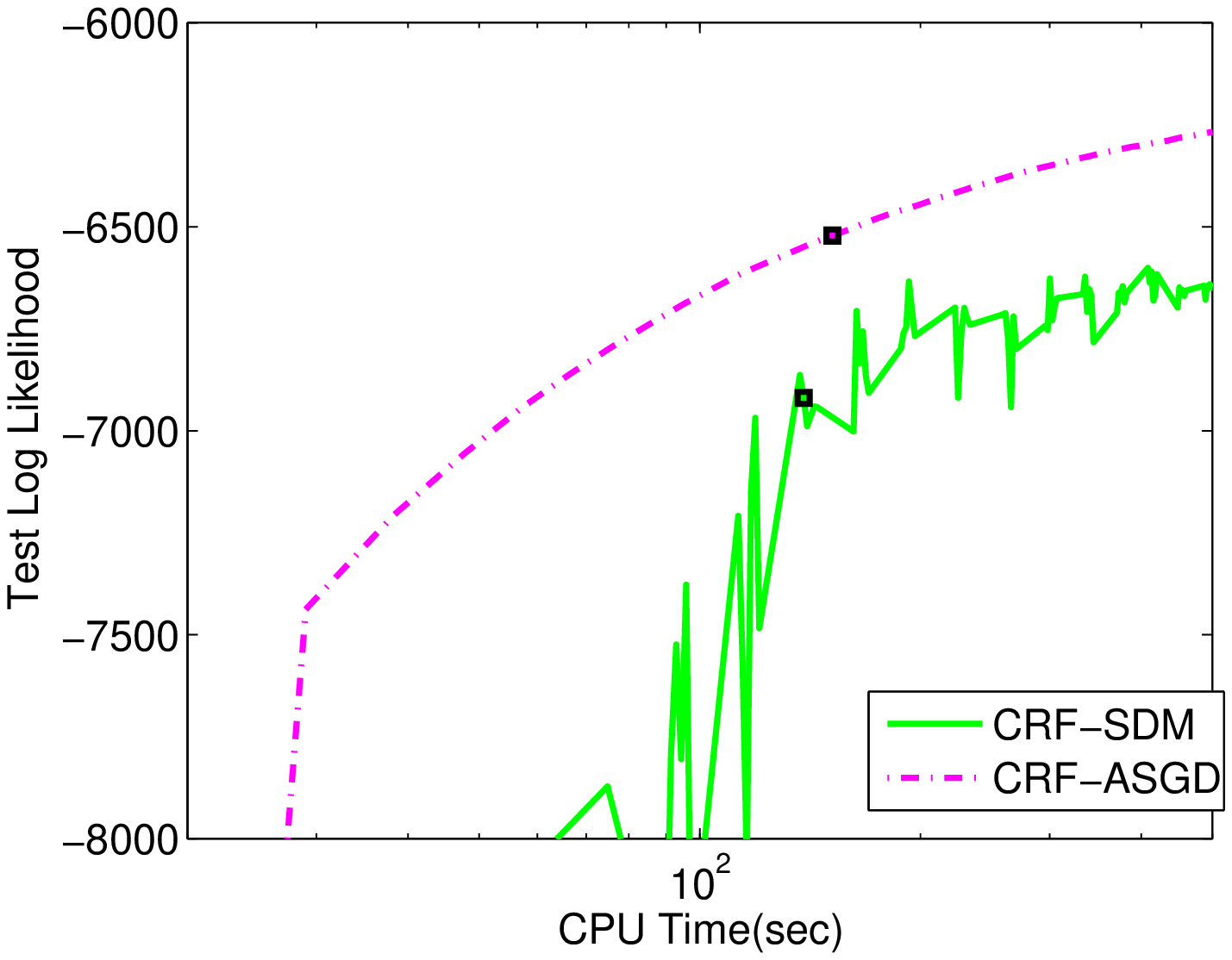}
\end{minipage}
\begin{minipage}[b]{0.5\linewidth}
\centering
\includegraphics[scale=0.4]{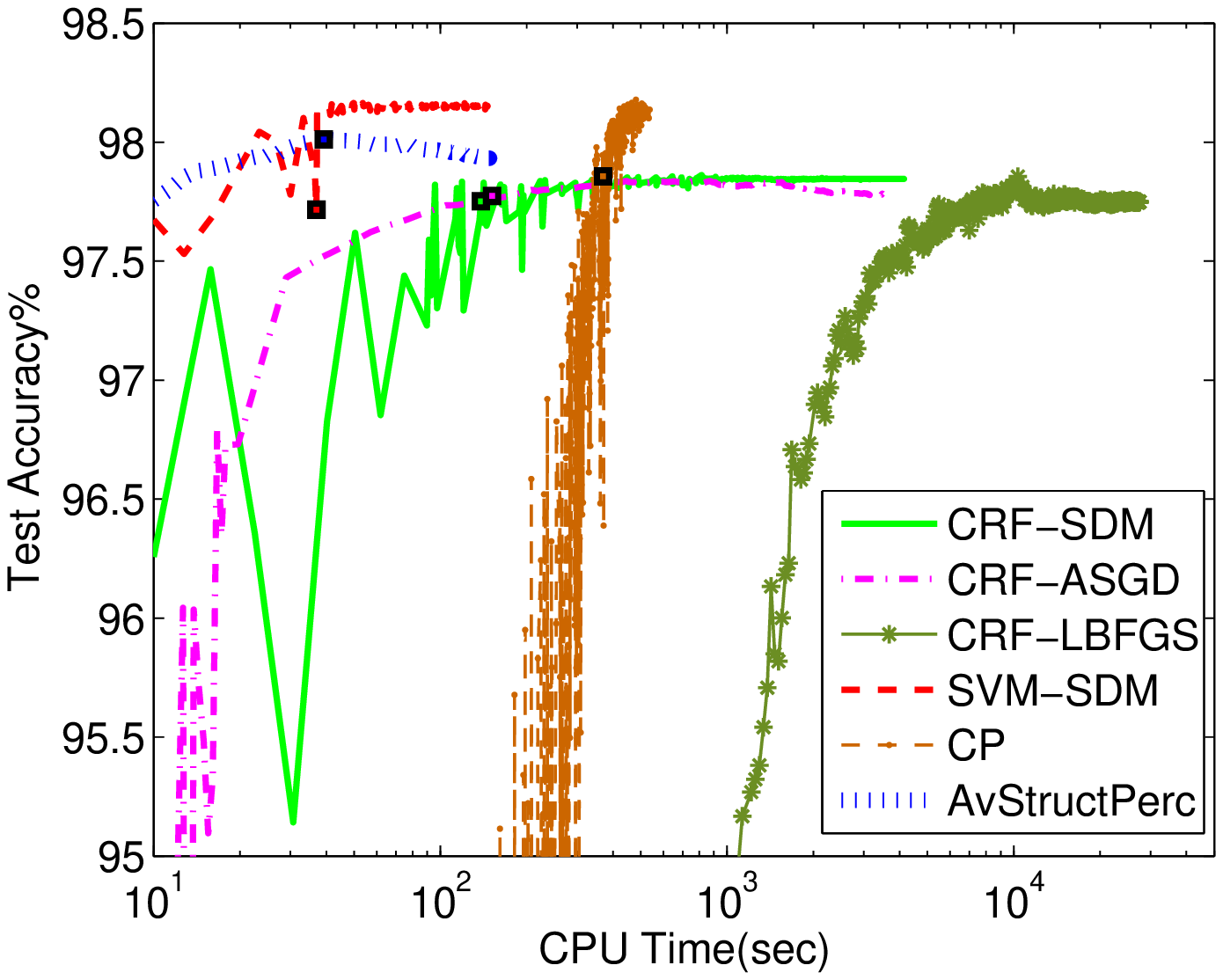}
\end{minipage}
\begin{minipage}[b]{0.5\linewidth}
\centering
\includegraphics[scale=0.4]{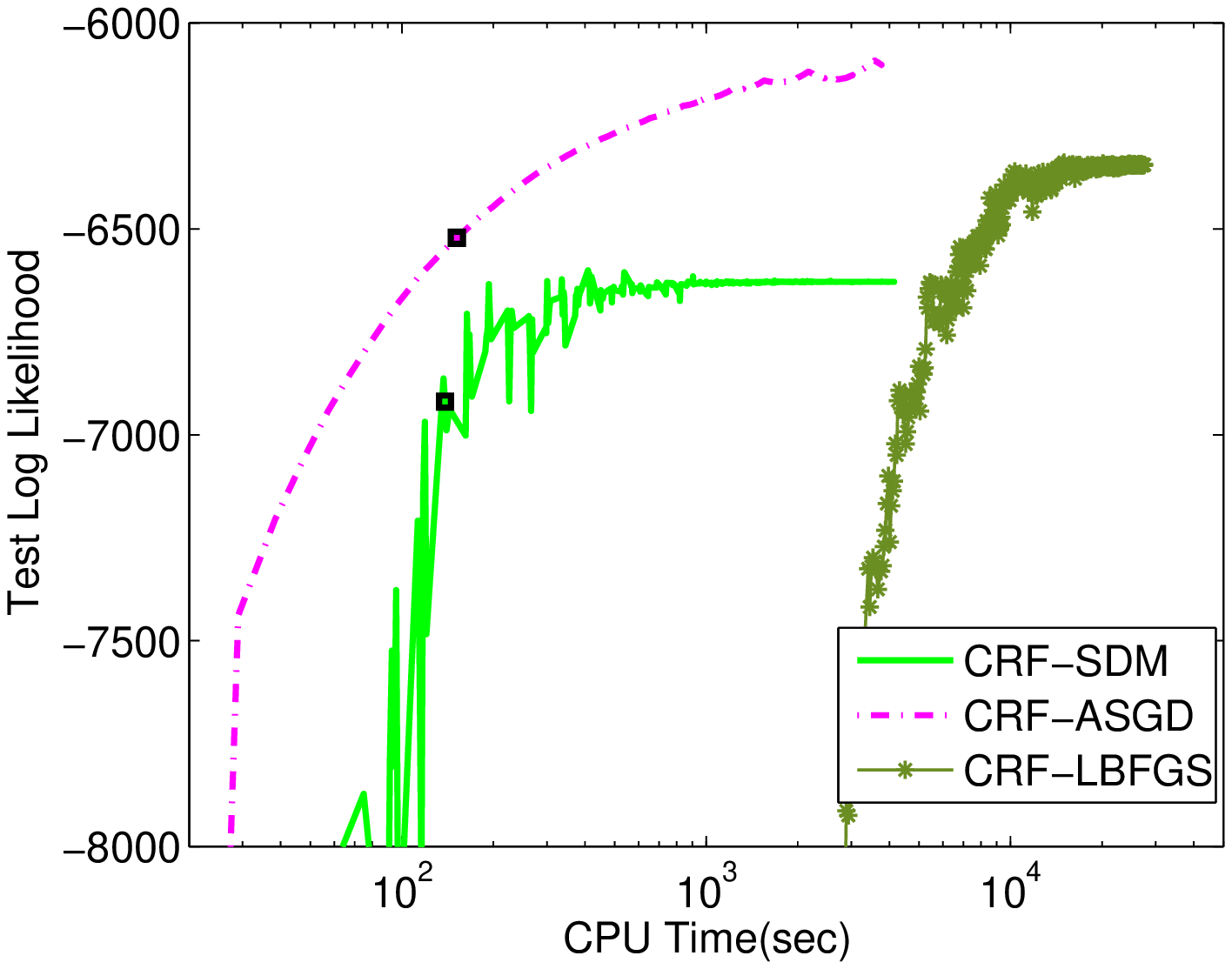}
\end{minipage}
\caption{\textbf{Comparison of Test Accuracy and Test Likelihood for BioNLP dataset.The plots in rows 2 and 3 are zoomed versions to clearly see certain
behaviour in the initial and final stages respectively.} }
\label{bionlpfig}
\end{figure}

\begin{figure*}[!h]
\begin{minipage}[b]{0.5\linewidth}
\centering
\includegraphics[scale=0.4]{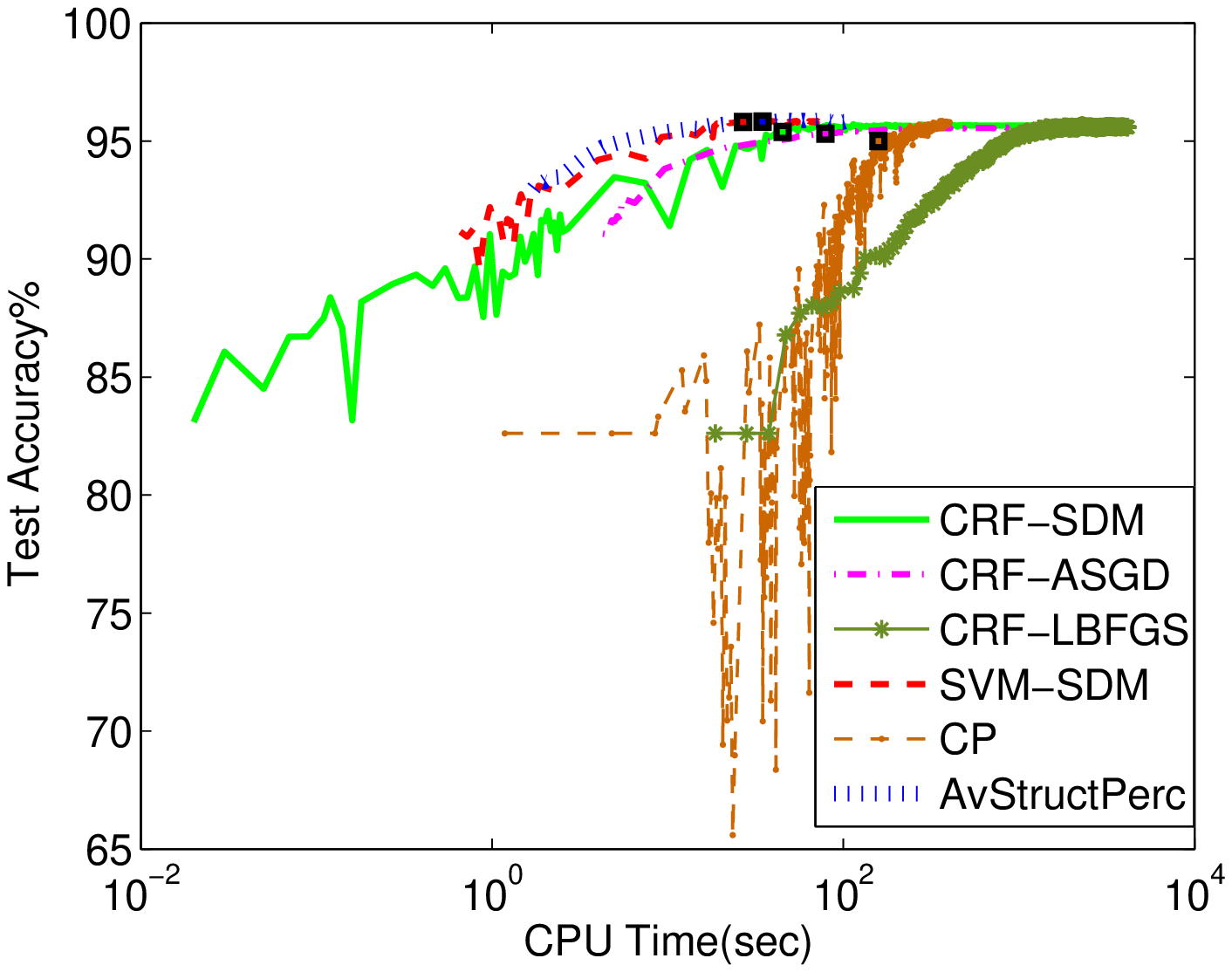}
\end{minipage}
\begin{minipage}[b]{0.5\linewidth}
\centering
\includegraphics[scale=0.4]{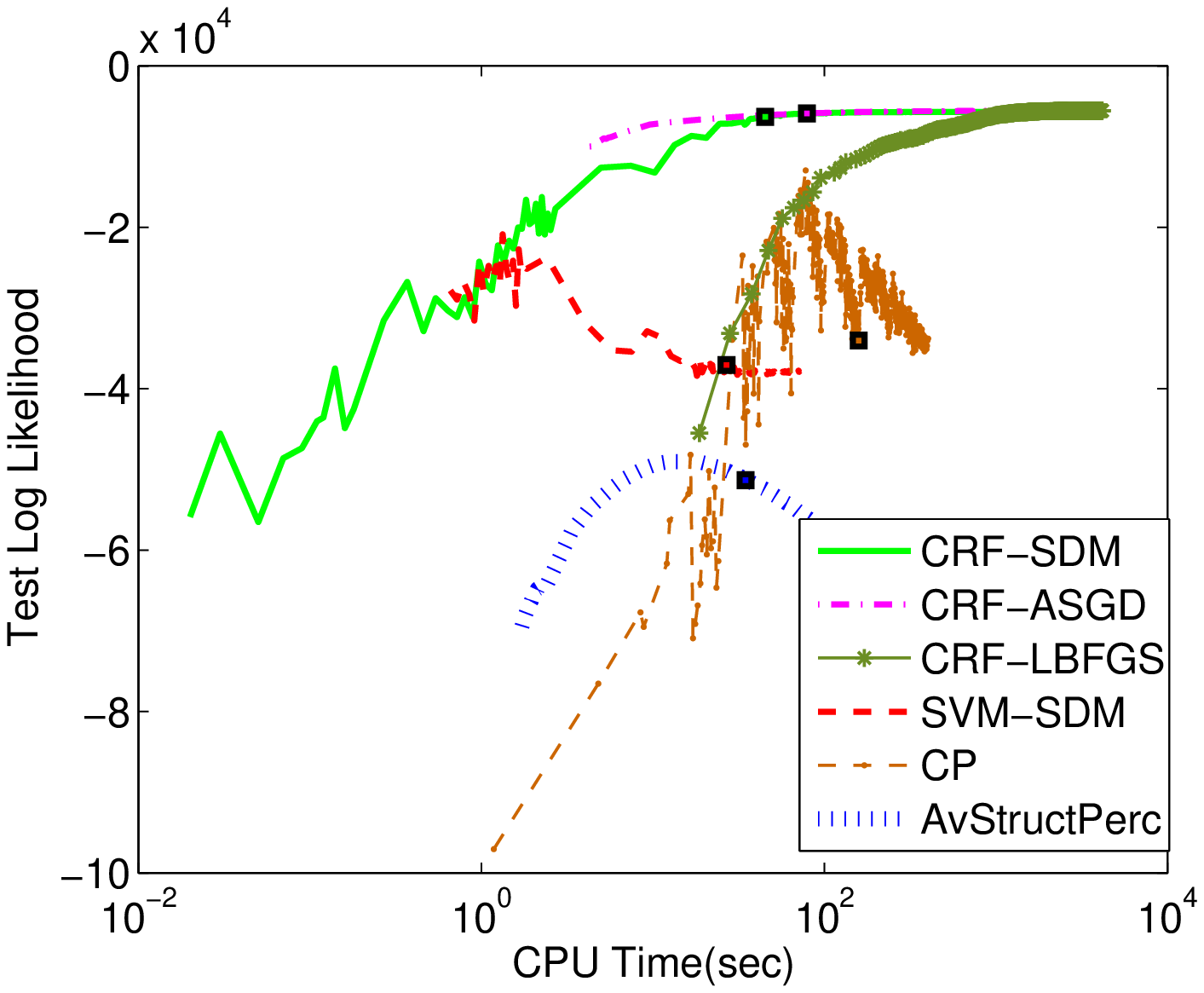}
\end{minipage}
\begin{minipage}[b]{0.5\linewidth}
\centering
\includegraphics[scale=0.4]{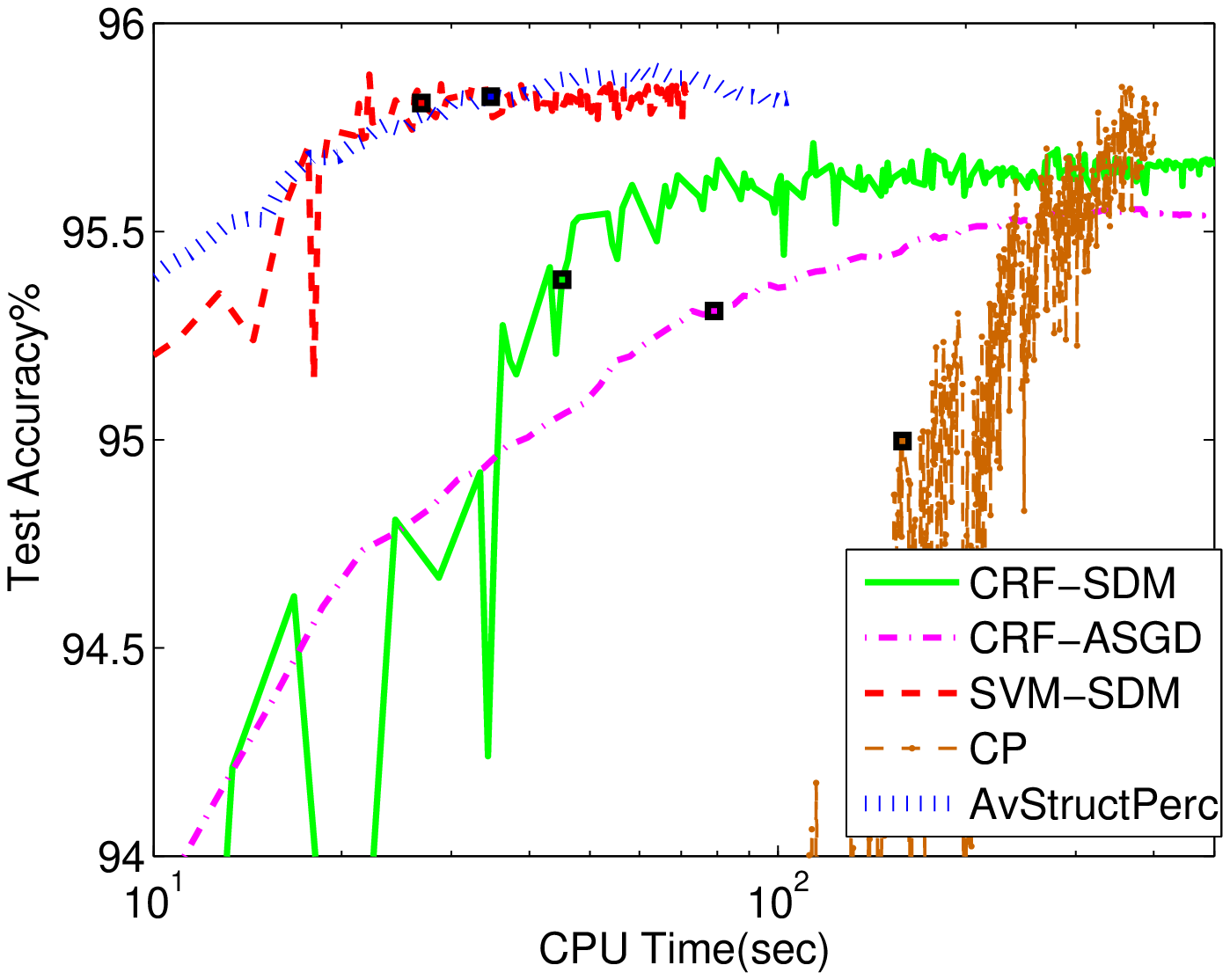}
\end{minipage}
\begin{minipage}[b]{0.5\linewidth}
\centering
\includegraphics[scale=0.4]{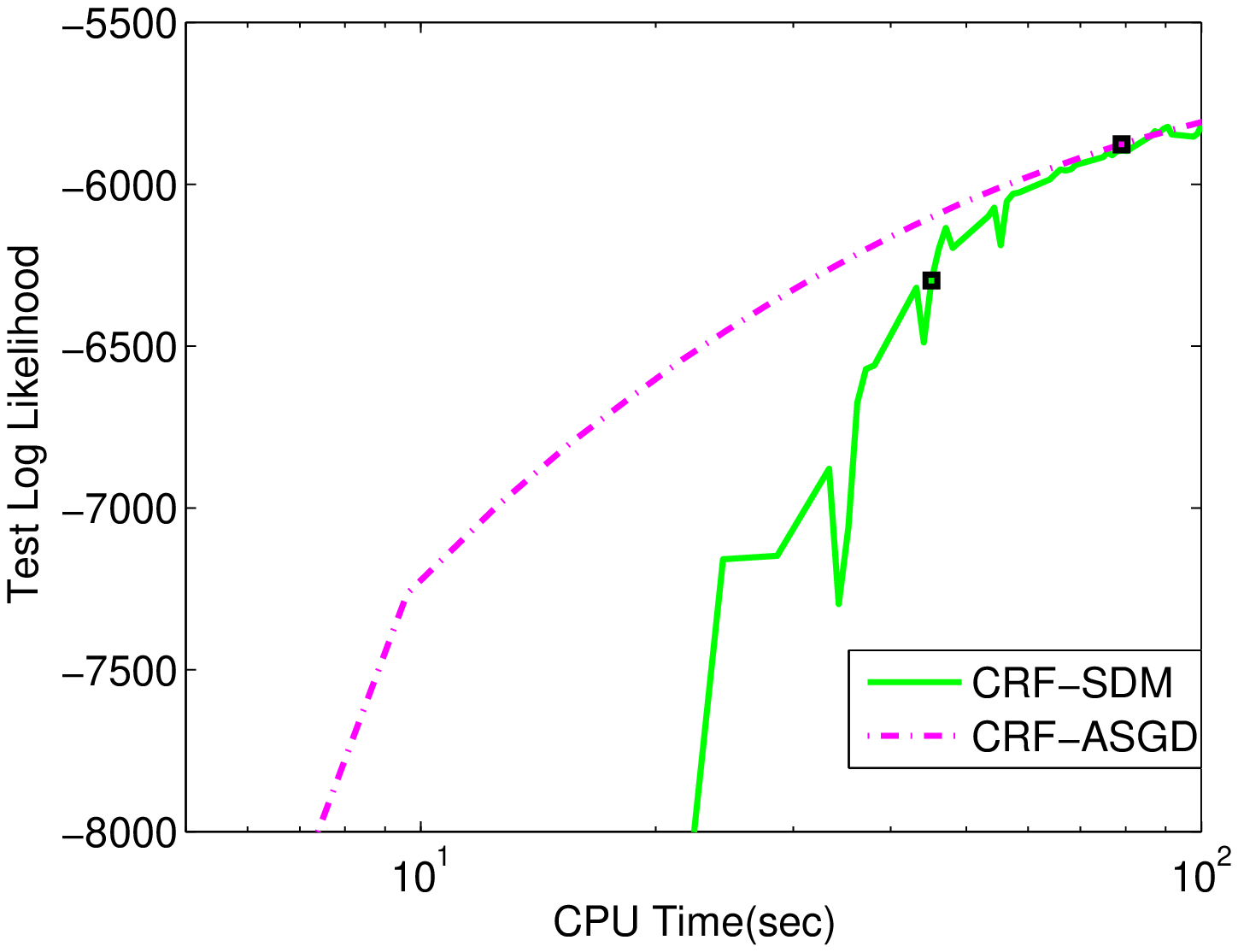}
\end{minipage}
\begin{minipage}[b]{0.5\linewidth}
\centering
\includegraphics[scale=0.4]{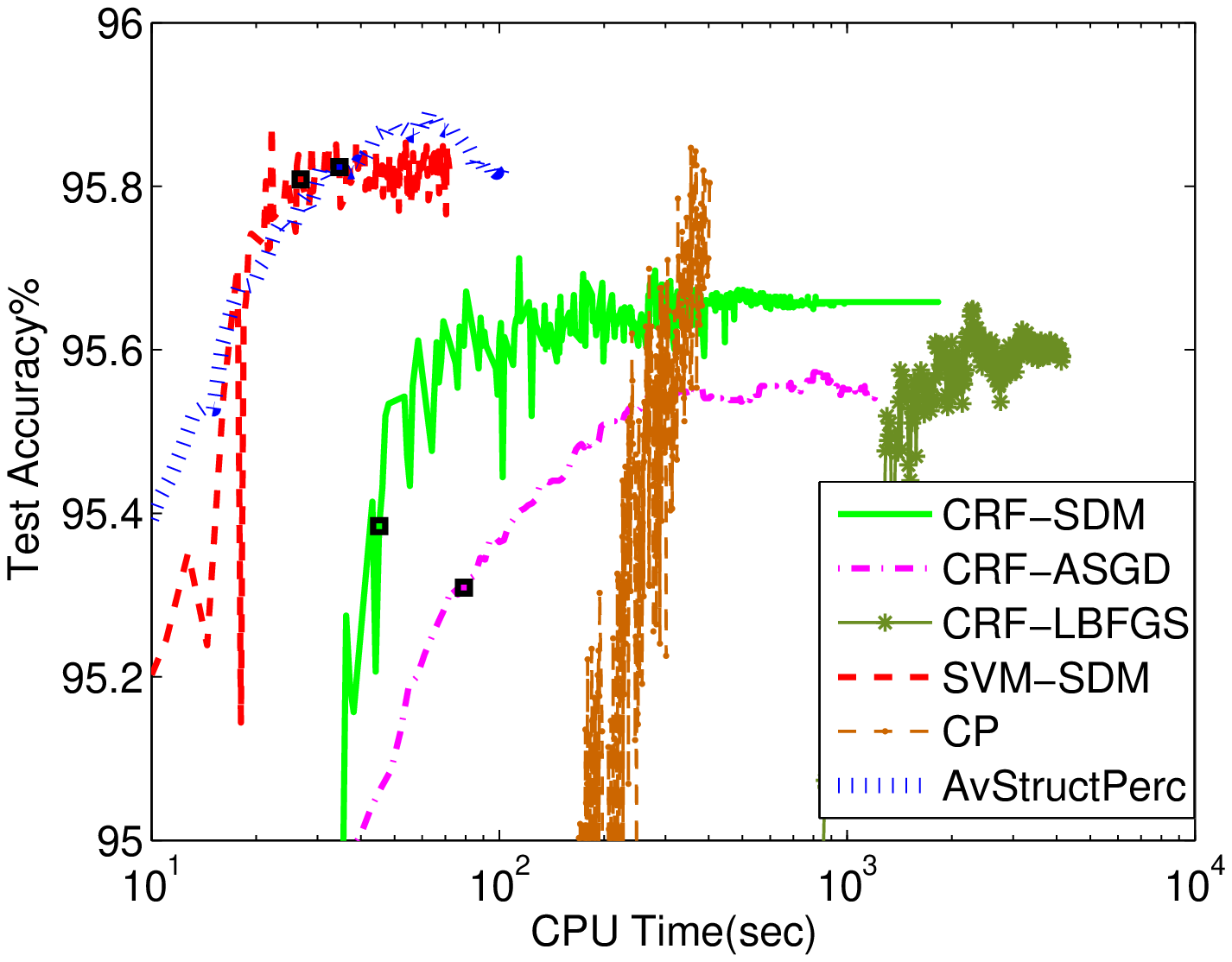}
\end{minipage}
\begin{minipage}[b]{0.5\linewidth}
\centering
\includegraphics[scale=0.4]{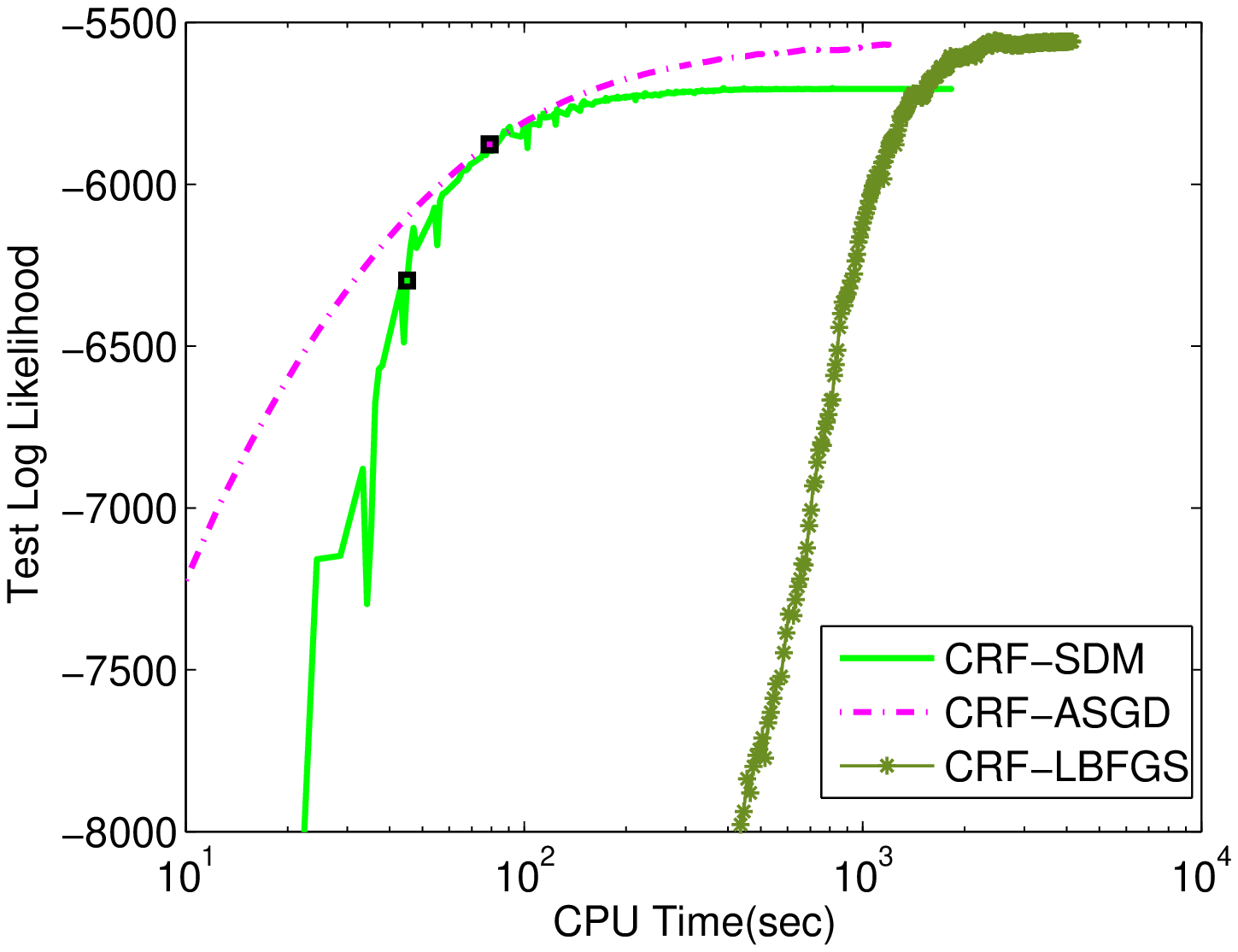}
\end{minipage}
\caption{\textbf{Comparison of Test Accuracy and Test Likelihood for dataCoNLL2003 dataset.
 The plots in rows 2 and 3 are zoomed versions to clearly see certain
behaviour in the initial and final stages respectively.
 } }
\label{dataconll2003fig}
\end{figure*}

\begin{figure*}[!h]
\begin{minipage}[b]{0.5\linewidth}
\centering
\includegraphics[scale=0.4]{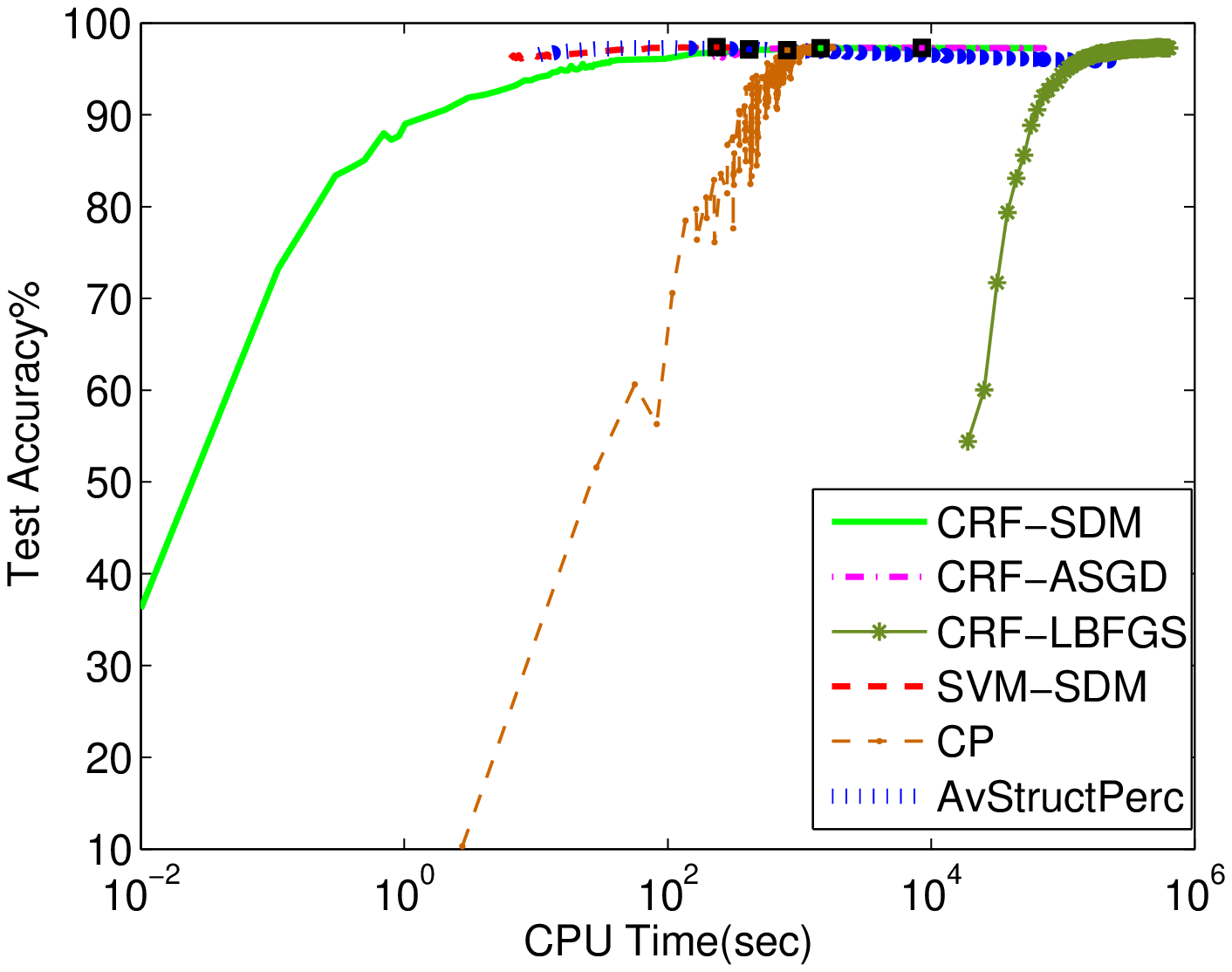}
\end{minipage}
\begin{minipage}[b]{0.5\linewidth}
\centering
\includegraphics[scale=0.4]{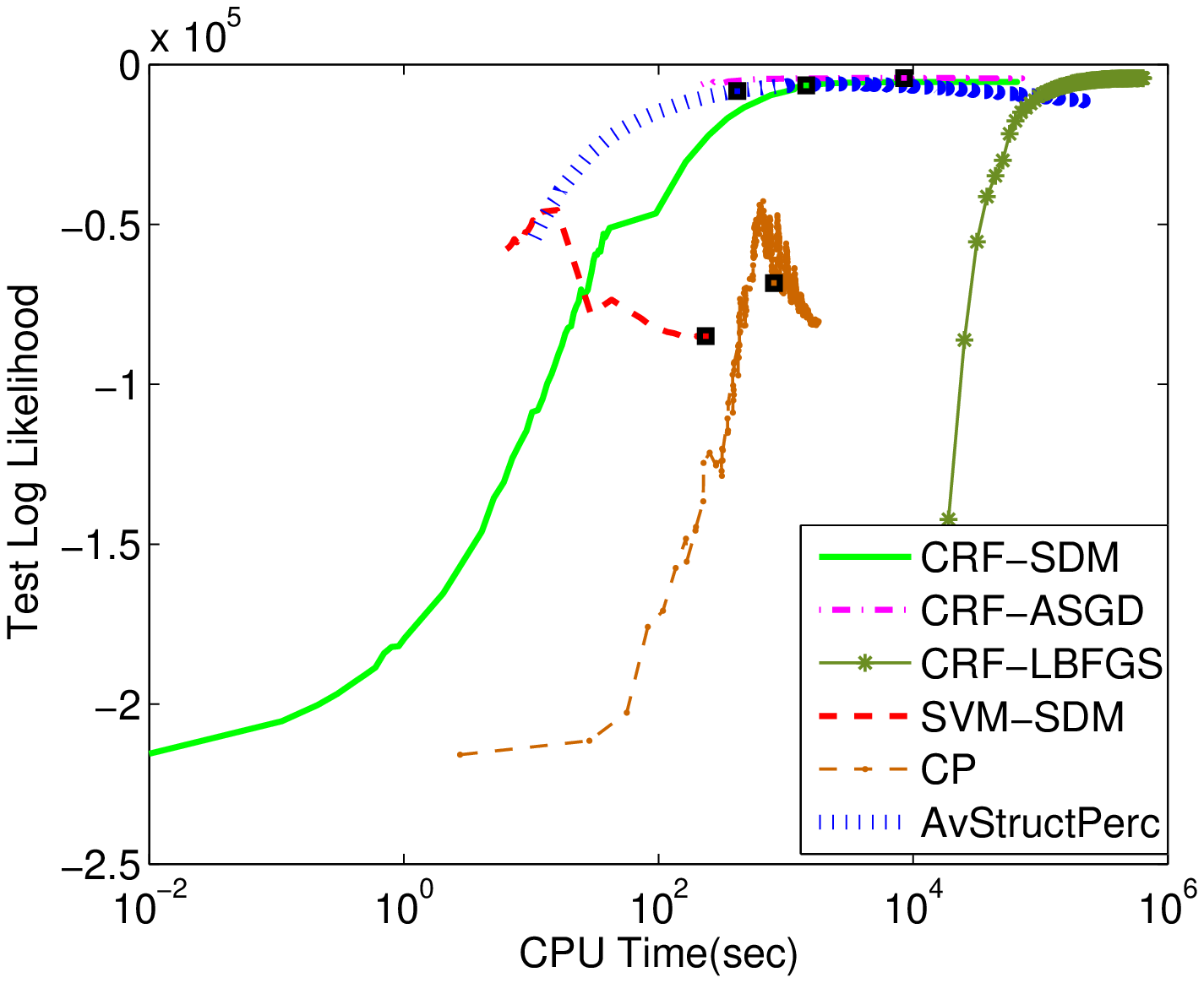}
\end{minipage}
\begin{minipage}[b]{0.5\linewidth}
\centering
\includegraphics[scale=0.4]{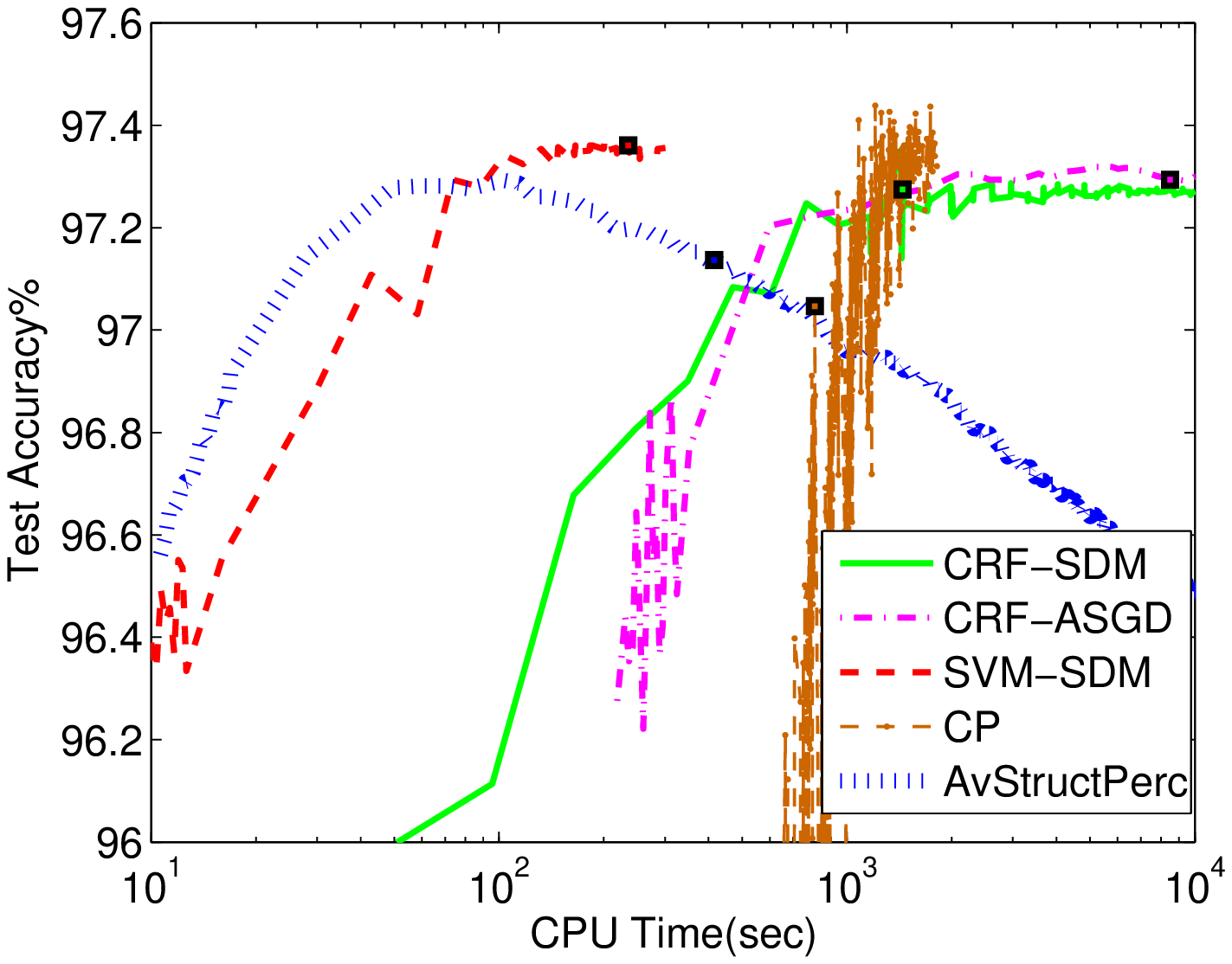}
\end{minipage}
\begin{minipage}[b]{0.5\linewidth}
\centering
\includegraphics[scale=0.4]{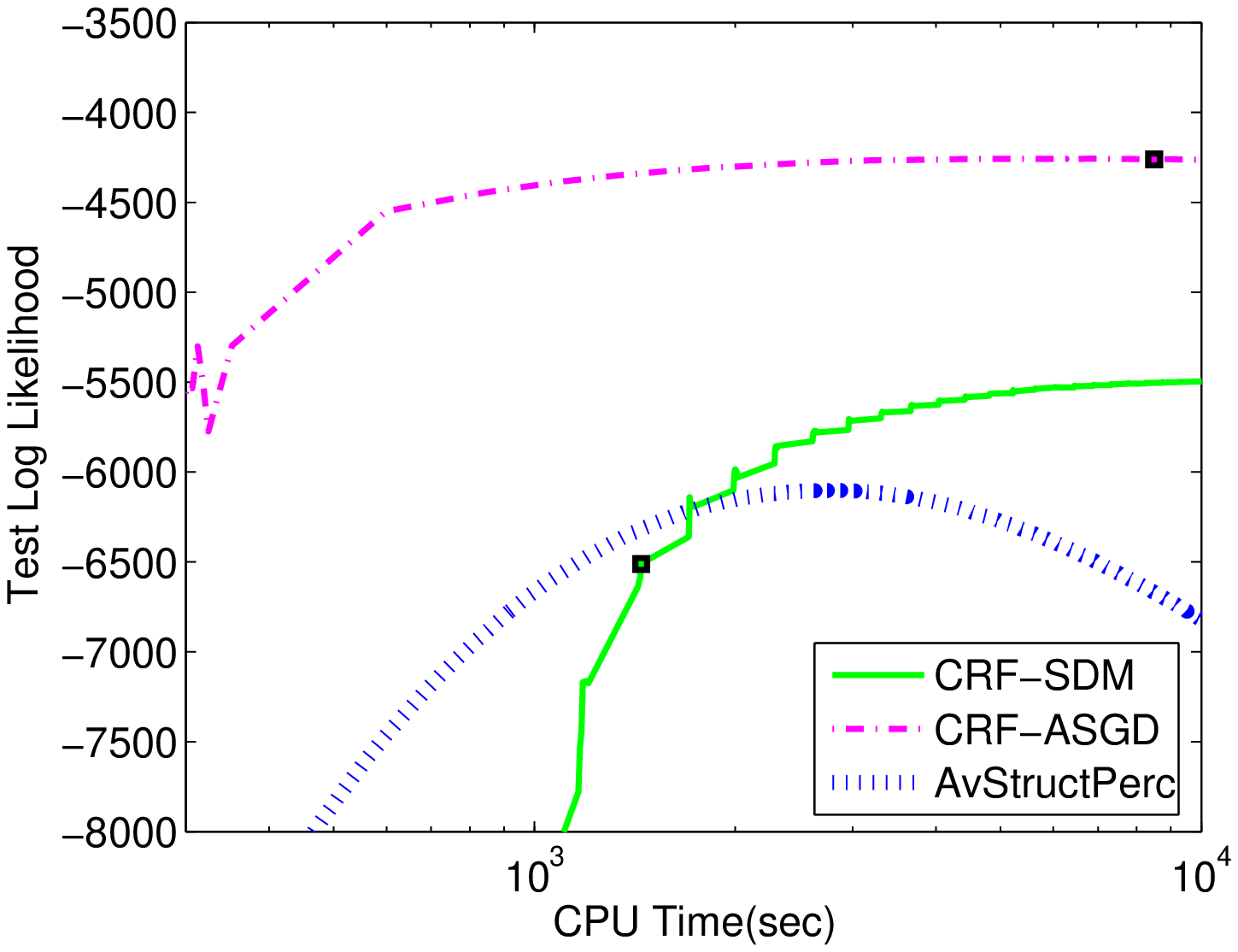}
\end{minipage}
\begin{minipage}[b]{0.5\linewidth}
\centering
\includegraphics[scale=0.4]{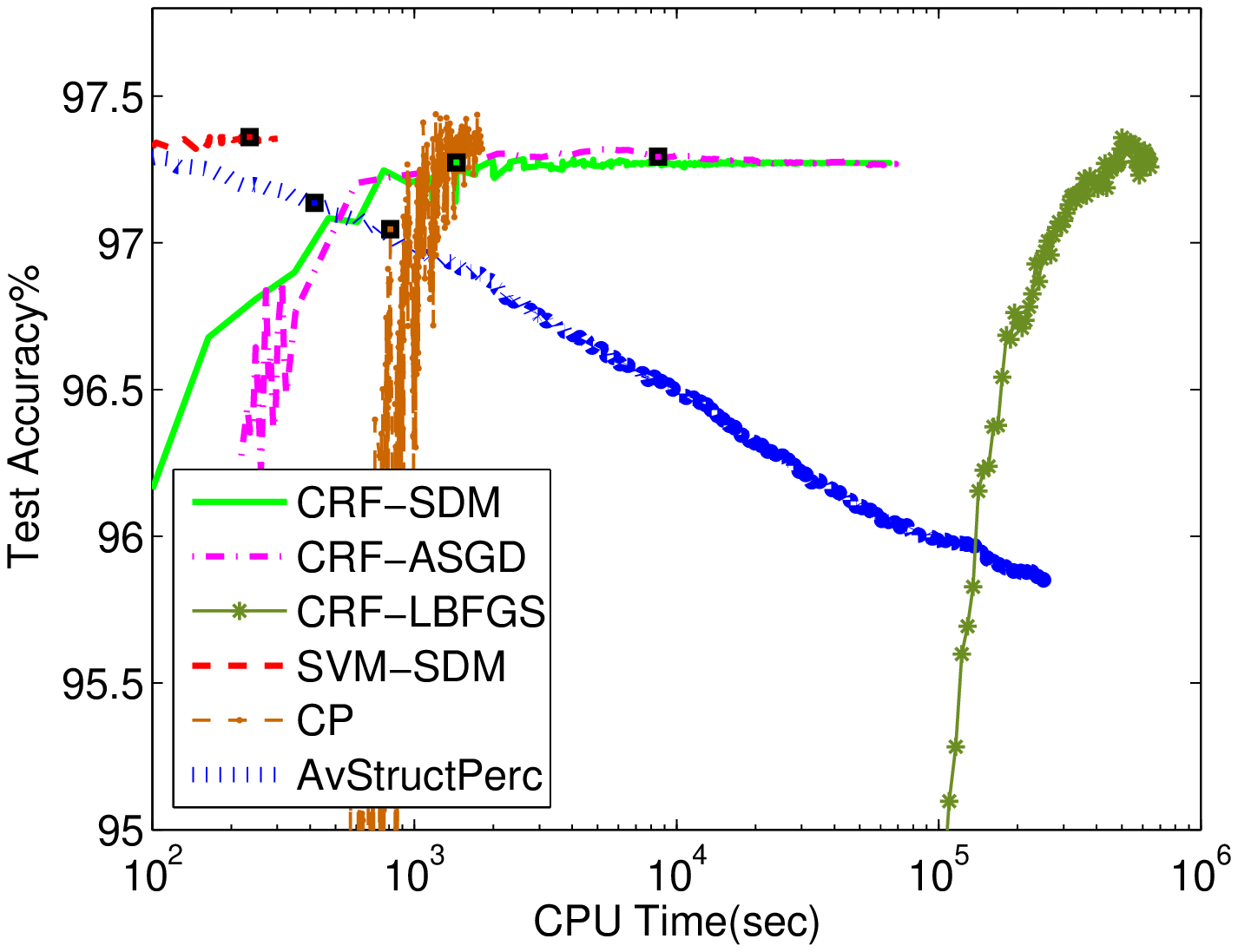}
\end{minipage}
\begin{minipage}[b]{0.5\linewidth}
\centering
\includegraphics[scale=0.4]{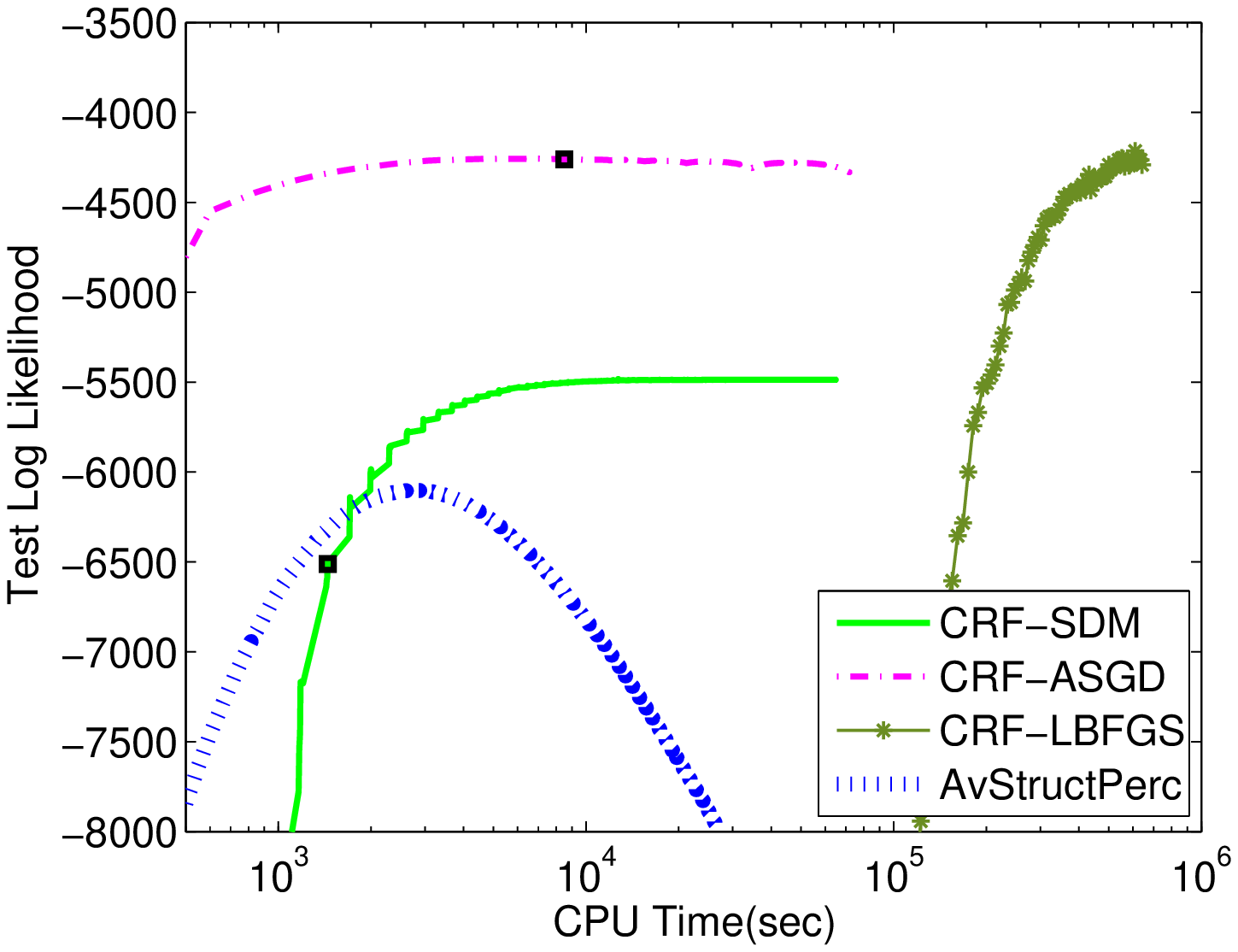}
\end{minipage}
\caption{\textbf{Comparison of Test Accuracy and Test Likelihood for dataPOS dataset.
 The plots in rows 2 and 3 are zoomed versions to clearly see certain
behaviour in the initial and final stages respectively.
} }
\label{dataposfig}
\end{figure*}

\begin{figure*}[ht]
\begin{minipage}[b]{0.5\linewidth}
\centering
\includegraphics[scale=0.4]{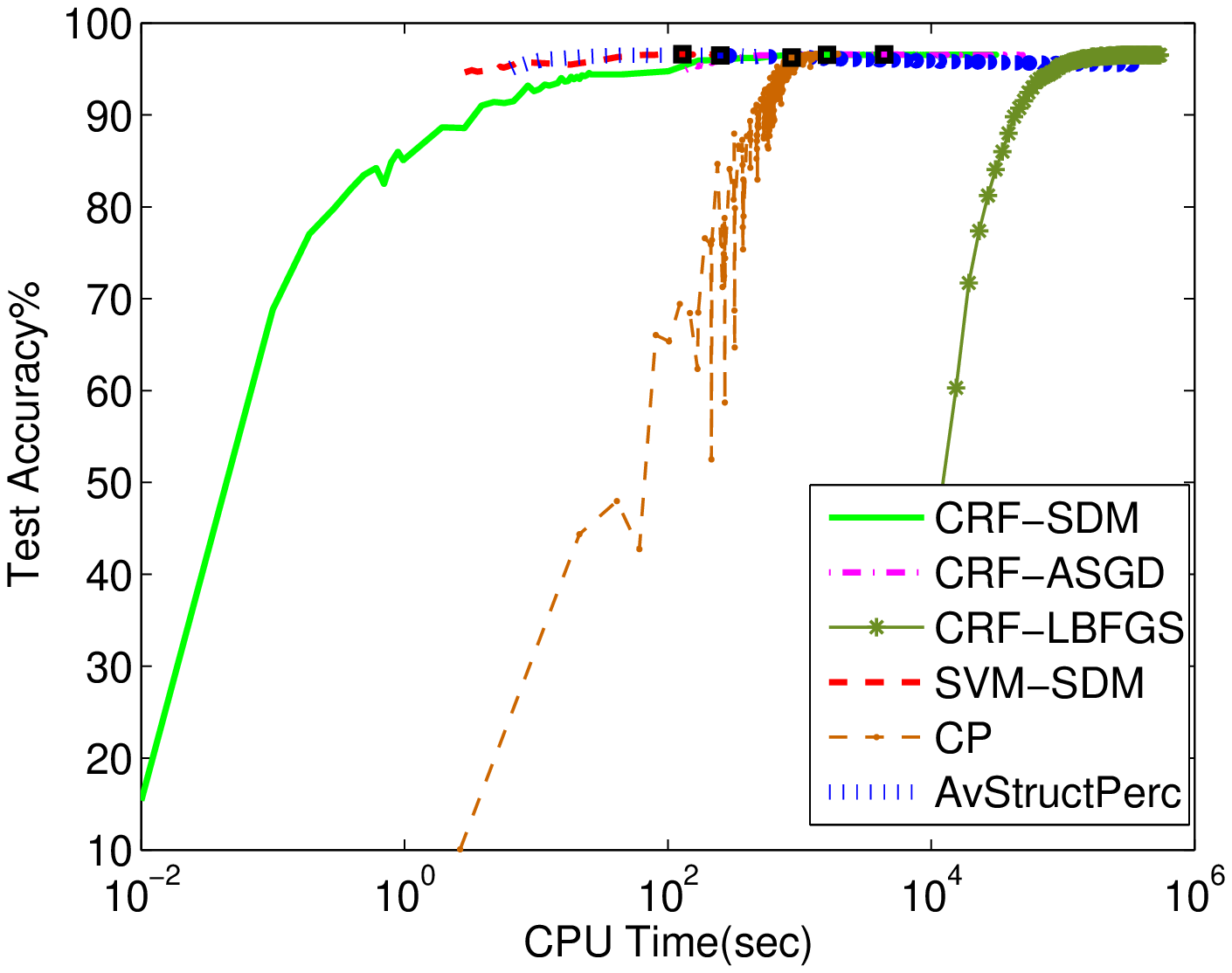}
\end{minipage}
\begin{minipage}[b]{0.5\linewidth}
\centering
\includegraphics[scale=0.4]{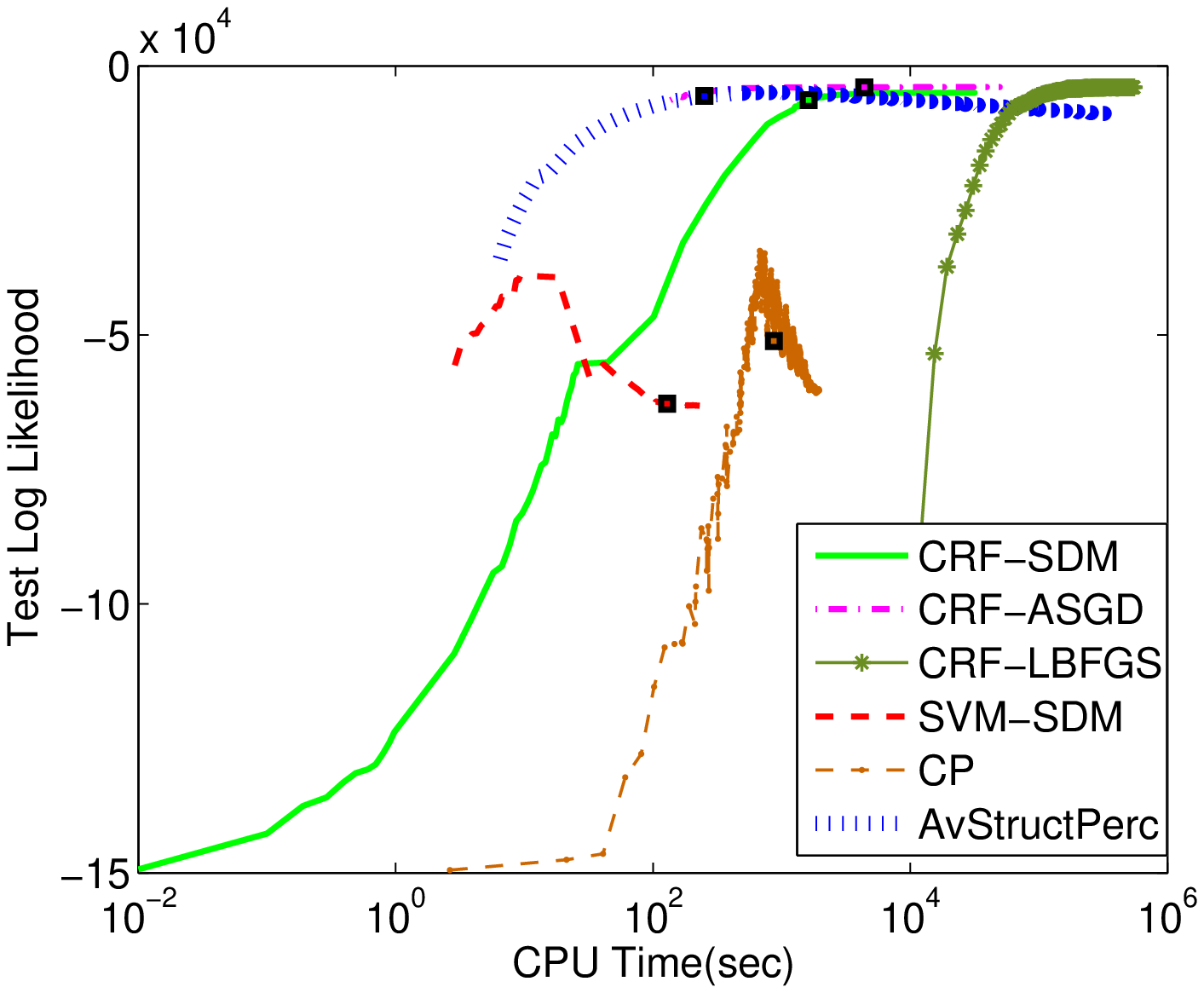}
\end{minipage}
\begin{minipage}[b]{0.5\linewidth}
\centering
\includegraphics[scale=0.4]{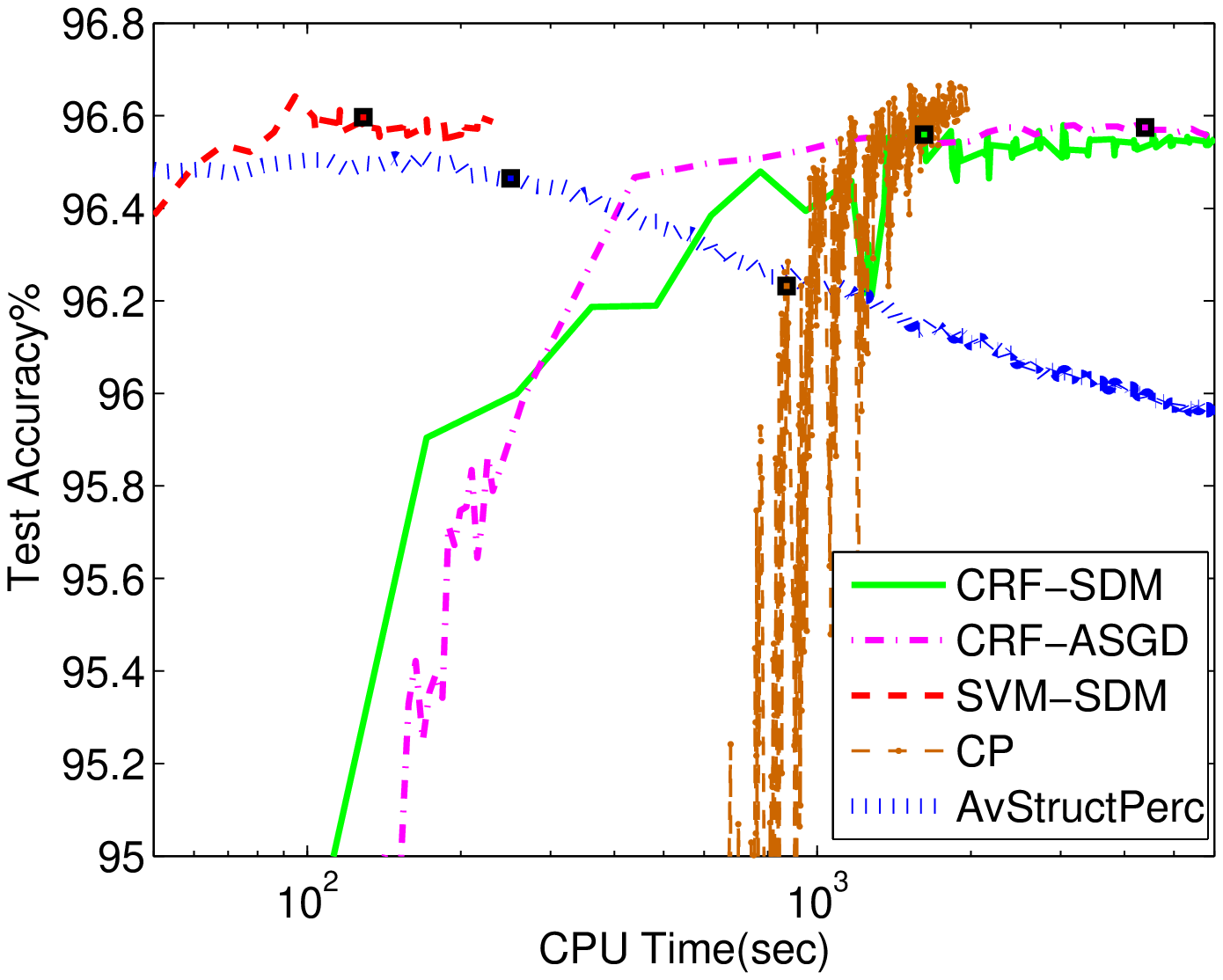}
\end{minipage}
\begin{minipage}[b]{0.5\linewidth}
\centering
\includegraphics[scale=0.4]{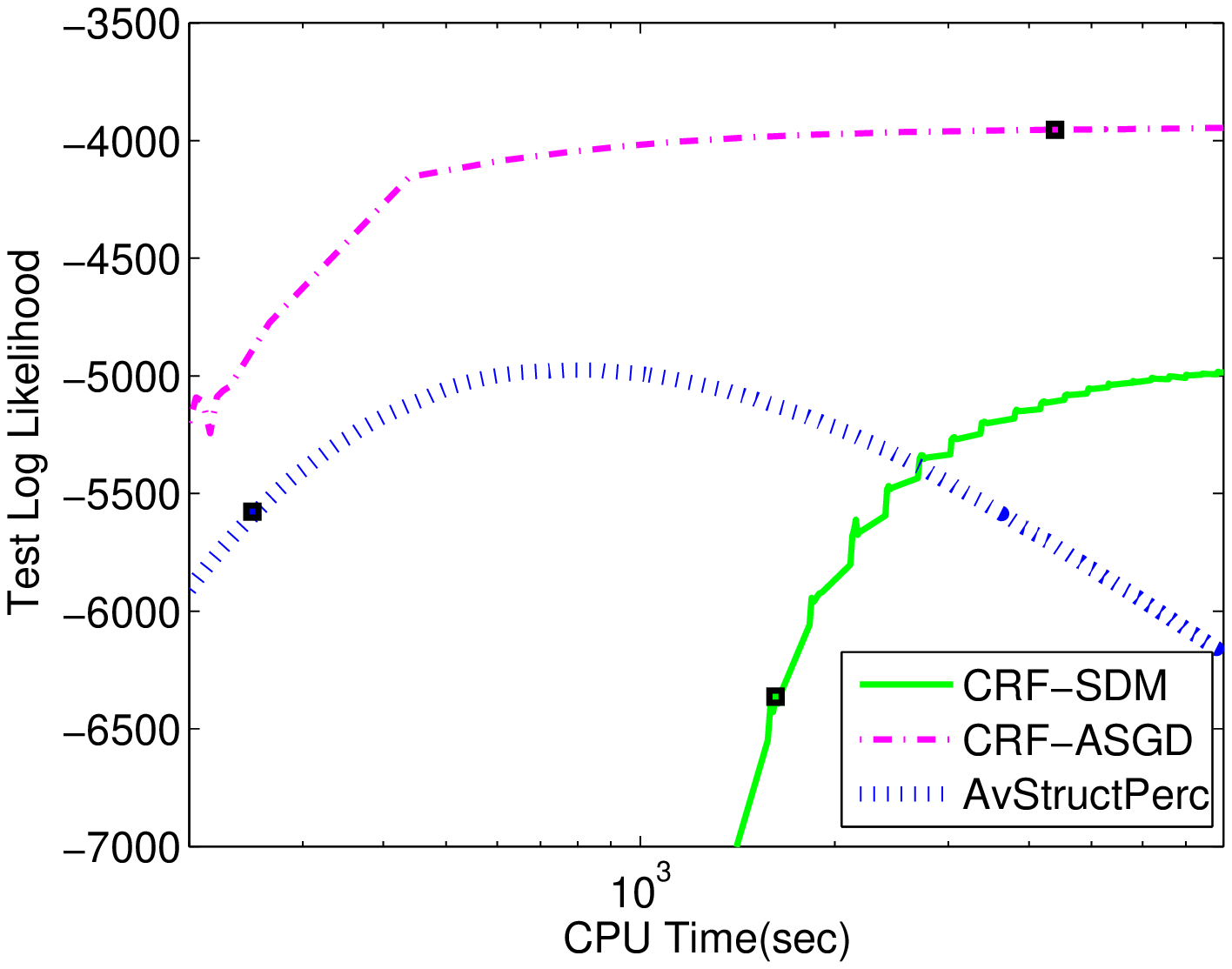}
\end{minipage}
\begin{minipage}[b]{0.5\linewidth}
\centering
\includegraphics[scale=0.4]{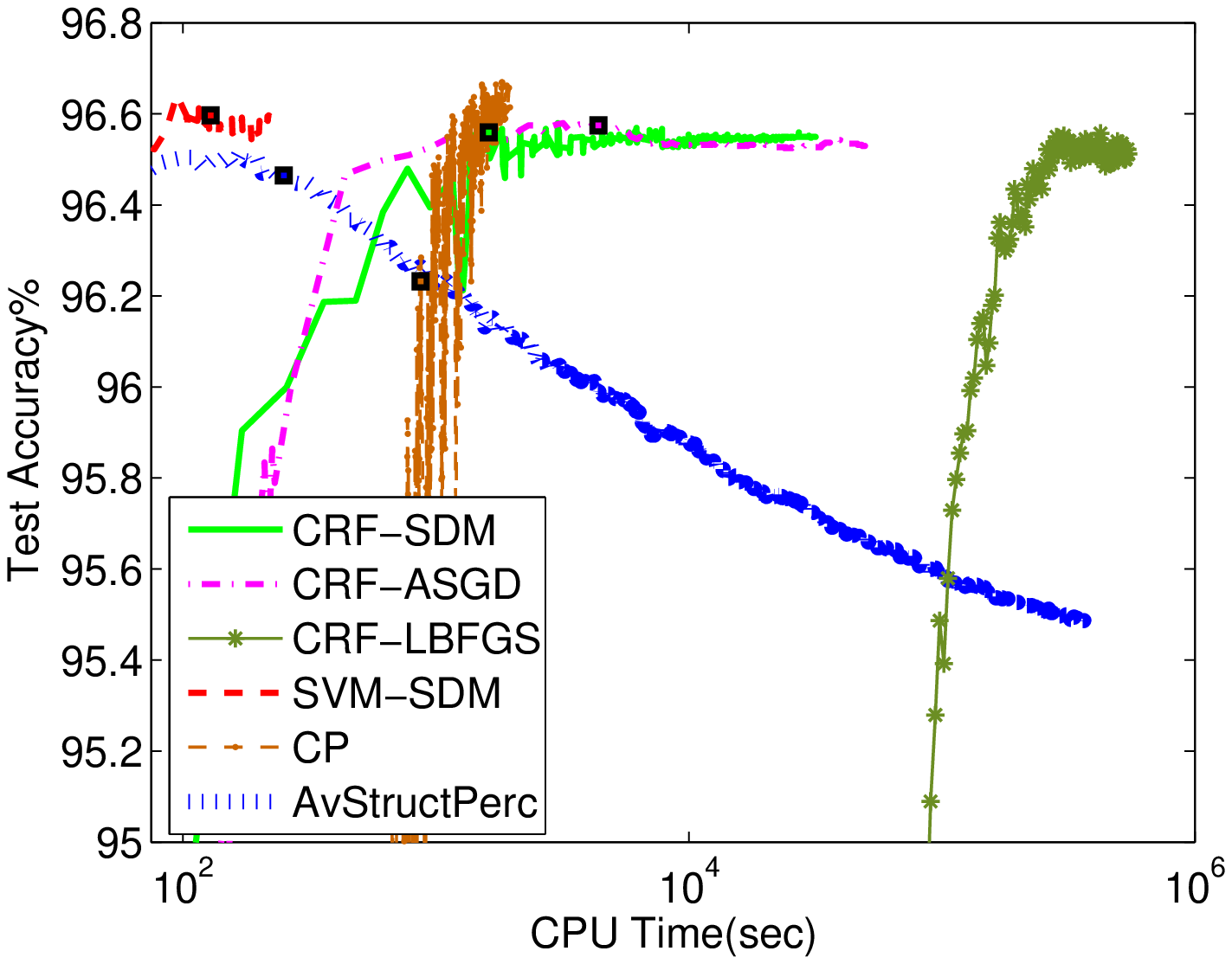}
\end{minipage}
\begin{minipage}[b]{0.5\linewidth}
\centering
\includegraphics[scale=0.4]{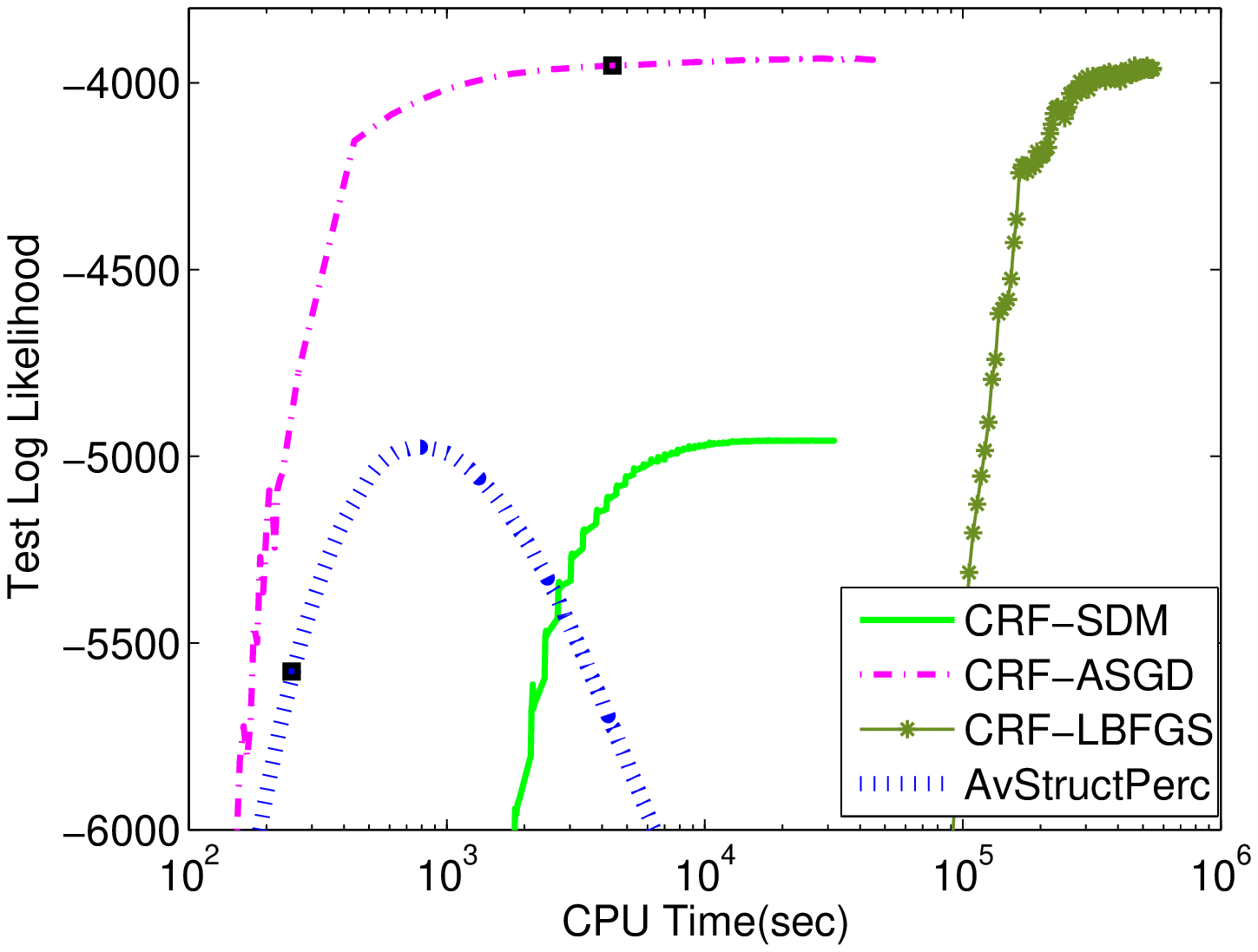}
\end{minipage}
\caption{\textbf{Comparison of Test Accuracy and Test Likelihood for WSJPOS dataset.
 The plots in rows 2 and 3 are zoomed versions to clearly see certain
behaviour in the initial and final stages respectively.
} }
\label{wsjposfig}
\end{figure*}

In this section we compare different sequence learning algorithms, mentioned
in Section~\ref{seqla}, on five benchmark data sets:
\begin{itemize}
 \item Wall Street Journal POS (WSJPOS)~\mycite{wsjpos}
 \item BioNLP~\mycite{bionlp}
 \item BioCreative~\mycite{biocre}
 \item CoNLL~\mycite{conll2003} and 
 \item dataPOS~\mycite{datapos}.
\end{itemize}
%
The characteristics of these data sets 
are summarized in Table~\ref{dataset_tab}. 
In all our experiments, the feature vector $f(\bx,\by)$ was 
constructed using the combination of first order and token-independent
second order feature functions~\mycite{crfcomp}. 
For the first order feature function, we aggregated the $d$ dimensional 
feature vector over the nodes for each label. The second order functions
used are independent of the tokens, but capture the dependencies between
the labels of two neighbouring nodes.
In this case, the dimension $N$ of the feature vector is $k^2+d k$.
The value of the hyper-parameter $\lambda$ was set to $10$ and $1$ respectively 
for the problems~(\ref{eq:mm-primal}) and~(\ref{eq:ll-primal}).
To compare the test likelihood achieved by batch CRF  with that obtained 
by the models trained using online algorithms, limited
memory BFGS method (CRF-LBFGS) was used in the batch mode. 
To get an idea about the steady state generalization performance of the 
resulting models trained using online algorithms, the algorithms were run for a large number of iterations.
Test set accuracy (for large-margin related methods) and likelihood (for CRF related methods) were used as performance measures to
compare different algorithms.
Validation set performance was calculated at the end of every iteration; this
time was not counted for CPU time calculations.
An algorithm could be stopped if there is no significant improvement in
the validation set performance.
This stopping criterion was used for all the methods.
This condition is represented  by a black square on each graph in Figures~\ref{biocreativefig}-\ref{wsjposfig}.
The computation of $\arg\max_{\by} \bw(\Alpha)^T \fiy$ and similar 
terms was  
performed using the Viterbi algorithm. All experiments were run on a $2.4$GHz Intel Xeon  
processor with $16$GB of shared main memory under Linux.
The plots in Figures~\ref{biocreativefig}-\ref{wsjposfig}. depict the performance of different algorithms. 
The results are summarized in Table~\ref{largedatacompare}.

{\bf Generalization Performance (Test Set Accuracy)} : The left 
panel of Figures~\ref{biocreativefig}-\ref{wsjposfig}. shows the behaviour of test set accuracy of
the resulting models, as a function of CPU time. From these plots, it is 
evident that the SVM-SDM method achieves the steady state performance much faster
than other methods. The difference is significant especially 
for large data sets like dataPOS and WSJPOS (Figure~\ref{dataposfig} and Figure~\ref{wsjposfig}, left panel). It is worth
noting that the performance of the averaged structured perceptron is comparable with 
SVM-SDM on BioCreative and CoNLL data sets(Figure~\ref{biocreativefig} and Figure~\ref{dataconll2003fig}). 
On large data sets, however, it showed degrading behaviour
as iterations progressed (Figures~\ref{dataposfig}-\ref{wsjposfig}).
The averaged structured perceptron has this overfitting/overtraining issue
that is bothersome. The methods which solve the
problem~(\ref{eq:ll-primal}) or its dual were not found to be suitable
if test set accuracy is used as a performance measure. The main reason is 
that these methods are designed to maximize the likelihood. On the other hand,
methods like CRF-ASGD perform better if test set likelihood is used 
as a performance measure. We now discuss this.

\begin{table}[!h]
    \caption{\textbf{Comparison of various algorithms at the point of termination based on validation set performance. Acc - Accuracy, ANLL - Average Negative Log-likelihood. The best and the second best results are highlighted in boldface and italic style respectively.}}
    \label{largedatacompare}
    \begin{center}
    \small
    \begin{tabular}{|c|c|c|c|c|}
    \hline %
             Dataset & Algorithm & Time (sec) & \small{Test Acc(\%)} & Test ANLL   \\\hline
	    \small{BioCreative}	&	\small{AvStructPerc}	&	{\bf 4.84}	&	98.67	&	7800  \\\hline
		      
	    		&	\small{SVM-SDM}	& {\it 7.21} & {\bf 98.73} & 21820  \\\hline

	    		&	\small{CP}	& 31.17 & {\it 98.72} & 16420  \\\hline

	    		&	\small{CRF-SDM}	& 20.95 & 98.6 & {\em 5053}  \\\hline

	    		&	\small{CRF-ASGD}	& 15.16 & 98.38 & 5491  \\\hline

      	    		&	\small{CRF-LBFGS}		& 270.8 & 98.58 & {\bf 4667}  \\\hline\hline
	    BioNLP	&	\small{AvStructPerc}	&	{\it 39.23}	&	{\bf 98.01}	&	210600  \\\hline
		      
	    		&	\small{SVM-SDM}	& {\bf 36.93} & 97.72 & 56170  \\\hline

	    		&	\small{CP}	& 369.7 & {\it 97.86} & 28730  \\\hline

	    		&	\small{CRF-SDM}	& 138.5 & 97.75 & 6920  \\\hline

	    		&	\small{CRF-ASGD}	& 151.7 & 97.77 & {\it 6522}  \\\hline

      	    		&	\small{CRF-LBFGS}		& 8850 & 97.76 & {\bf 6439}  \\\hline\hline
%
%
%
%
	   CoNLL	&	\small{AvStructPerc}	&	{\it 34.69}	&	{\bf 95.8}	&	51330  \\\hline
		      
	    		&	\small{SVM-SDM}	& {\bf 26.85} & {\bf 95.8} & 37060  \\\hline

	    		&	\small{CP}	& 158.4 & 95 & 34020  \\\hline

	    		&	\small{CRF-SDM}	& 45.11 & 95.38 & 6297 \\\hline

	    		&	\small{CRF-ASGD}	& 79.02 & 95.31 & {\it 5877}  \\\hline

      	    		&	\small{CRF-LBFGS}		& 1350 & 95.51 & {\bf 5765}  \\\hline\hline
	   dataPOS	&	\small{AvStructPerc}	&	{\it 415.3}	&	97.14	&	8256  \\\hline
		      
	    		&	\small{SVM-SDM}	& {\bf 235.2} & {\bf 97.36} & 84900  \\\hline

	    		&	\small{CP}	& 808.7 & 97.05 & 68310  \\\hline

	    		&	\small{CRF-SDM}	& 1446 & 97.27 & 6511 \\\hline

	    		&	\small{CRF-ASGD}	& 8497 & {\it 97.29} & {\it 4261}  \\\hline

      	    		&	\small{CRF-LBFGS}		& 609600 & 97.28 & {\bf 4215}  \\\hline\hline

	   WSJPOS	&	\small{AvStructPerc}	&	{\it 250.6}	&	96.46	&	5576  \\\hline
		      
	    		&	\small{SVM-SDM}	& {\bf 128.8} & {\bf 96.6} & 627900  \\\hline

	    		&	\small{CP}	& 870.7 & 96.23 & 511200  \\\hline

	    		&	\small{CRF-SDM}	& 1620 & 96.56 & 6363\\\hline

	    		&	\small{CRF-ASGD}	& 4391 & {\it 96.57} & {\it 3954}  \\\hline

      	    		&	\small{CRF-LBFGS}		& 463700 & 96.5 & {\bf 3953}  \\\hline\hline

    \end{tabular}
    \end{center}
\end{table}

{\bf Generalization Performance (Test Set  Likelihood)} : The behaviour 
of test set likelihood as a function of CPU time, is
depicted in the right panel of Figures~\ref{biocreativefig}-\ref{wsjposfig}.
Note that the methods
CRF-ASGD and CRF-LBFGS (batch algorithm) optimize the training set 
likelihood directly by solving problem~(\ref{eq:ll-primal}). 
On the other hand, the sequential dual method for CRFs (CRF-SDM)
solves the dual problem of~(\ref{eq:ll-primal}). 
From the results in Table~\ref{largedatacompare}, we see that   
the CRF-LBFGS method, though slow, gave the best test likelihood value 
among all
the methods compared.
While the performance of CRF-SDM and CRF-ASGD was comparable on
three data sets,
the CRF-ASGD method clearly outperformed the CRF-SDM method on
large data sets like dataPOS and WSJPOS. This
is mainly due to the large number of alphabets ($k$ in Table 1)
in these data sets. This results in a large number of possible output 
sequences for every example. 
Since the CRF-SDM method assumes that, only those sequences in the set 
$V_i$ have a non-zero value of the dual variable, and the set $V_i$ 
cannot accommodate all possible sequences associated with example $i$, the
vector ${\bw}(\Alpha)$ in~(\ref{eq:wdual}) cannot be accurately determined.
This results in degradation of performance of CRF-SDM, especially on those
data sets where the number of alphabets is large.

Test likelihood performance of CP and SVM-SDM methods was not good 
compared to other methods, as these methods were mainly designed to solve the dual
problems obtained using a variant of the hinge loss function.
On the other hand, the averaged structured perceptron, which does not optimize any objective function, performed better than these methods on all the data sets except BioNLP and CoNLL. 
The averaged structured perceptron algorithm is simple, easy to implement, fast and gives reasonable test accuracy and likelihood performance. 
However, it does begin to overfit eventually, as is evident from Figures~\ref{biocreativefig}-\ref{wsjposfig}.
Some key observations, made in this Section, are summarized in 
Table~\ref{perfallmethods}.

{\bf Sensitivity to hyper-parameter selection}: For the Structural SVM and CRF model 
design, the effect of hyper-parameter selection on the test set performance was studied by conducting 
the following experiment. The hyper-parameter $\lambda$ (in \eqref{eq:mm-primal} or \eqref{eq:ll-primal}) 
was varied from $10^{-3}$ to $10^{3}$ and the test set accuracy and likelihood values were noted. 
The results are reported in Table~\ref{sensitivitytable}. 
It is clear from this table that the variation in test accuracy was not large for Structural SVM 
over different values of $\lambda$. On the other hand, test likelihood values 
varied significantly over a range of $\lambda$ values, when CRF was used. 
We however note that, for most of the data sets, $\lambda=10$ was an optimal choice for 
Structural SVMs (with accuracy as a measure) and the corresponding choice for CRF was $\lambda=1$ (with likelihood 
as a measure). These optimal choices of hyper-parameters were used in all our experiments.

\section{ Conclusion and Recommendations}
\label{conclusion}
In this work, we have done a systematic comparison of different 
sequence labeling algorithms in terms of their speed,
ability to reach a good generalization performance (accuracy and 
likelihood) fast, and 
ability to maintain best generalization performance at the end. 
Based on
experimental results on real-world benchmark data sets,
we recommend that a dual method like SVM-SDM which solves the dual 
of~(\ref{eq:mm-primal}) is preferred if test accuracy is the evaluation
 criterion. 
The averaged structured perceptron yields good test accuracy; however it has
an overfitting issue that is bothersome. 
On the other
hand, the CRF-ASGD method should be preferred if likelihood is the criterion.

 \begin{table}[ht]
    \caption{\textbf{Sensitivity of Structural SVM and CRF to} $\l$. \textbf{ANLL - Average Negative Log-likelihood. (Not given for Structural SVM as one does not expect any pattern there.) The best results are highlighted in boldface style.} }
    \label{sensitivitytable}
    \begin{center}
    \small
    \begin{tabular}{|c|c|c|c|c|}
    \hline %
             Dataset & $\l$ & \multicolumn{2}{c|}{Test Accuracy (\%)} & Test ANLL\\\cline{3-5}
		      & & Struct SVM & CRF & CRF \\
		      & &   (SDM)    & (ASGD) & (ASGD)\\\hline
	  \small{BioCreative}&$10^{3}$	& 97.27 & 94.17 & 15186 \\\hline
	  &$10^{2}$	& 98.07 & 97.01 & 9278 \\\hline
	  &10 		& 98.76 & 98.16 & 6070 \\\hline
	  &1 		& {\bf 98.77} & 98.5 & 4894 \\\hline
	  &$10^{-1}$	& 98.36 & 98.57 & 4725 \\\hline
	  &$10^{-2}$	& 98.07 & 98.57 & {\bf 4707}  \\\hline
	  &$10^{-3}$	& 97.47 & 98.57 & 4707 \\\hline\hline
	  \small{BioNLP}&$10^{3}$	& 96.46 & 91.6 & 21105 \\\hline
	  &$10^{2}$	& 97.83 & 96.07 & 10752 \\\hline
	  &10 		& {\bf 98.15} & 97.45 & 7115 \\\hline
	  &1 		& 98.02 & 97.82 & {\bf 6189} \\\hline
	  &$10^{-1}$ 	& 97.56 & 97.88 & 6452 \\\hline
	  &$10^{-2}$	& 97.29 & 97.88 & 6550 \\\hline
	  &$10^{-3}$	& 96.5 & 97.88 & 6553 \\\hline\hline
	  \small{CoNLL}&$10^{3}$	& 91.75 & 89.57 & 12763 \\\hline
	  &$10^{2}$	& 93.94 & 92.01 & 9076 \\\hline
	  &10 		& {\bf 95.81} & 94.6 & 6680 \\\hline
	  &1		& 95.6 & 95.54 & {\bf 5630} \\\hline
	  &$10^{-1}$	& 94.75 & 95.74 & 5659 \\\hline
	  &$10^{-2}$       & 94.03 & 95.75 & 5715 \\\hline
	  &$10^{-3}$       & 92.59 & 95.75 & 5716 \\\hline\hline
	  \small{dataPOS}&$10^{3}$        & 95.1 & 90.7 & 25560 \\\hline
	  &$10^{2}$	& 96.7 & 95.79 & 9989 \\\hline
	  &10 		& {\bf 97.36} & 97.05 & 5358 \\\hline
	  &1 		& 97.04 & 97.27 & {\bf 4273} \\\hline
	  &$10^{-1}$	& 96.17 & 97.13 & 5142 \\\hline
	  &$10^{-2}$	& 95.01 & 96.99 & 6670 \\\hline
	  &$10^{-3}$	& 94.5  & 96.99 & 6770 \\\hline\hline
	  \small{WSJPOS}&$10^{3}$	& 93.4 & 87.87 & 21964 \\\hline
	  &$10^{2}$	& 95.88 & 94.36 & 8933  \\\hline
	  &10 		& {\bf 96.59} & 96.31 & 4769 \\\hline
	  &1 		& 96.25 & 96.53 & {\bf 3939} \\\hline
	  &$10^{-1}$	& 95.56 & 96.42 & 4544 \\\hline
	  &$10^{-2}$ 	& 95.01 & 96.35 & 5159 \\\hline
	  &$10^{-3}$ 	& 93.7 & 96.35 & 5188 \\\hline\hline
    \end{tabular}
    \end{center}
\end{table}

\begin{table*}[h]
     \caption{\textbf{Generalization performance characteristics of various algorithms}}
     \noindent
     \footnotesize
     \label{perfallmethods}
     \begin{center}
     \begin{tabular}{|p{3.3cm}|c|c|c|c|c|c|}
     \hline %
 	      & \multicolumn{6}{|c|}{Algorithm}\\
 	      \cline{2-7}
 	      &  AvgStrPerc & SVM-SDM & CP & CRF-SDM & CRF-SGD & CRF-LBFGS \\\hline 
 	      Ability to reach best test set & \multirow{2}{*}{\checkmark} & \multirow{2}{*}{\checkmark} & \multirow{2}{*}{\xmark} & \multirow{2}{*}{\xmark} & \multirow{2}{*}{\xmark} & \multirow{2}{*}{\xmark} \\
 	      accuracy fast & & & & & & \\\hline
 	      Ability to reach best test set  & \multirow{2}{*}{\xmark} & \multirow{2}{*}{\xmark} & \multirow{2}{*}{\xmark} & \multirow{2}{*}{\xmark} & \multirow{2}{*}{\checkmark} & \multirow{2}{*}{\xmark} \\
 	      likelihood fast & & & & & & \\\hline
 	      Ability to maintain & \multirow{2}{*}{\xmark} & \multirow{2}{*}{\checkmark} & \multirow{2}{*}{\checkmark} & \multirow{2}{*}{\checkmark} & \multirow{2}{*}{\checkmark} & \multirow{2}{*}{\checkmark} \\
 	      best test set accuracy & & & & & & \\\hline
 	      Ability to maintain & \multirow{2}{*}{\xmark} & \multirow{2}{*}{\xmark} & \multirow{2}{*}{\xmark} & \multirow{2}{*}{\checkmark} & \multirow{2}{*}{\checkmark} & \multirow{2}{*}{\checkmark} \\
 	      best test set likelihood & & & & & & \\\hline 
     \end{tabular}
     \end{center}
     \vspace{-.1in}
     \end{table*}
     
\clearpage
\bibliographystyle{apalike}

\bibliography{seqlabel}
\end{document}